\documentclass{article}
\usepackage{ijcai24}

\usepackage{times}
\usepackage{soul}
\usepackage{url}
\usepackage[hidelinks]{hyperref}
\usepackage[utf8]{inputenc}
\usepackage[small]{caption}
\usepackage{graphicx}
\usepackage{amsmath}
\usepackage{amsthm}
\usepackage{booktabs}
\usepackage{algorithm}
\usepackage{algorithmic}
\usepackage[switch]{lineno}

\usepackage{booktabs} 
\usepackage{tabulary,multirow,overpic,xcolor}
\usepackage{multirow}
\usepackage{pifont}
\usepackage{amsfonts}
\usepackage{subcaption}
\usepackage{listings}

\usepackage{colortbl}
\usepackage{enumitem}
\usepackage{braket}
\usepackage{array}

\usepackage{float}
\usepackage{bbding}
\usepackage{etoolbox} 

\usepackage{arydshln}
\usepackage{tabularx}
\usepackage{dsfont}

\usepackage{txfonts}

\definecolor{mygray}{gray}{.92}

\definecolor{imcolor}{RGB}{34, 139, 34}

\definecolor{frenchblue}{rgb}{0.0, 0.45, 0.73}

\newcommand{\gray}[1]{\textcolor{gray}{#1}}
\newcommand{\green}[1]{\textcolor[RGB]{96,187,87}{#1}}

\newcommand{\fn}[1]{\footnotesize{#1}}
\newcommand{\ybf}[1]{\gray{\bf{\fn{(#1)}}}}
\newcommand{\rbf}[1]{\color{purple}{\bf{\fn{(#1)}}}}
\newcommand{\gbf}[1]{\green{\bf{\fn{(#1)}}}}

\definecolor{cred}{HTML}{A10035}
\definecolor{cyellow}{HTML}{FEC260}
\definecolor{cgreen}{HTML}{3FA796}
\definecolor{cpurple}{HTML}{2A0944}
\definecolor{ggray}{RGB}{127,127,127}
\definecolor{aliceblue}{rgb}{0.94, 0.97, 1.0}
\definecolor{convnext_purple}{HTML}{483A7F}
\definecolor{convnext_yellow}{HTML}{E99675}

\newcommand{\npub}[1]{\color{gray}}

\definecolor{redcolor}{RGB}{255, 0, 0}
\newcommand{\textred}[1]{\textcolor{redcolor}{#1}}

\newcommand{\thickhline}{%
	\noalign {\ifnum 0=`}\fi \hrule height 1pt
	\futurelet \reserved@a \@xhline
}

\newcommand{\increase}[1]{
	\fontsize{6pt}{0.5em}\selectfont\green{$\uparrow$~\textbf{#1}}
}
\newcommand{\decrease}[1]{
	\fontsize{6pt}{0.5em}\selectfont\color{gray!48}{$\downarrow$~\textbf{#1}}
}
\newcommand{\wincrease}[1]{
	\fontsize{6pt}{0.5em}\selectfont\color{gray!48}{$\uparrow$~\textbf{#1}}
}

\newcolumntype{x}[1]{>{\centering\arraybackslash}p{#1pt}}
\newcolumntype{y}[1]{>{\raggedright\arraybackslash}p{#1pt}}
\newcolumntype{z}[1]{>{\raggedleft\arraybackslash}p{#1pt}}

\title{Self-Supervised Pre-training with Symmetric Superimposition Modeling for Scene
Text Recognition}

\author{
    Author Name
    \affiliations
    Affiliation
    \emails
    email@example.com
}

\author{
Zuan Gao
\and
Yuxin Wang\footnote{Corresponding Author}\and
Yadong Qu\and
Boqiang Zhang\and
\\
Zixiao Wang\and
Jianjun Xu\and
Hongtao Xie
\\
\affiliations
University of Science and Technology of China, Hefei, China\\
\emails
\{zuangao, qqqyd, cyril, wzx99, xujj1998\}@mail.ustc.edu.cn,
\{wangyx58, htxie\}@ustc.edu.cn
}

\begin{document}

\maketitle

\begin{abstract}
 In text recognition, self-supervised pre-training emerges as a good solution to reduce dependence on expansive annotated real data. Previous studies primarily focus on local visual representation by leveraging mask image modeling or sequence contrastive learning. However, they omit modeling the linguistic information in text images, which is crucial for recognizing text. To simultaneously capture local character features and linguistic information in visual space, we propose Symmetric Superimposition Modeling (SSM). The objective of SSM is to reconstruct the direction-specific pixel and feature signals from the symmetrically superimposed input. Specifically, we add the original image with its inverted views to create the symmetrically superimposed inputs. At the pixel level, we reconstruct the original and inverted images to capture character shapes and texture-level linguistic context. At the feature level, we reconstruct the feature of the same original image and inverted image with different augmentations to model the semantic-level linguistic context and the local character discrimination. In our design, we disrupt the character shape and linguistic rules. Consequently, the dual-level reconstruction facilitates understanding character shapes and linguistic information from the perspective of visual texture and feature semantics. Experiments on various text recognition benchmarks demonstrate the effectiveness and generality of SSM, with 4.1\% average performance gains and 86.6\% new state-of-the-art average word accuracy on Union14M benchmarks. The code is available at https://github.com/FaltingsA/SSM.

\end{abstract}
\vspace{-1.em}
\begin{figure}
    \centering 
\includegraphics[width=0.48\textwidth]{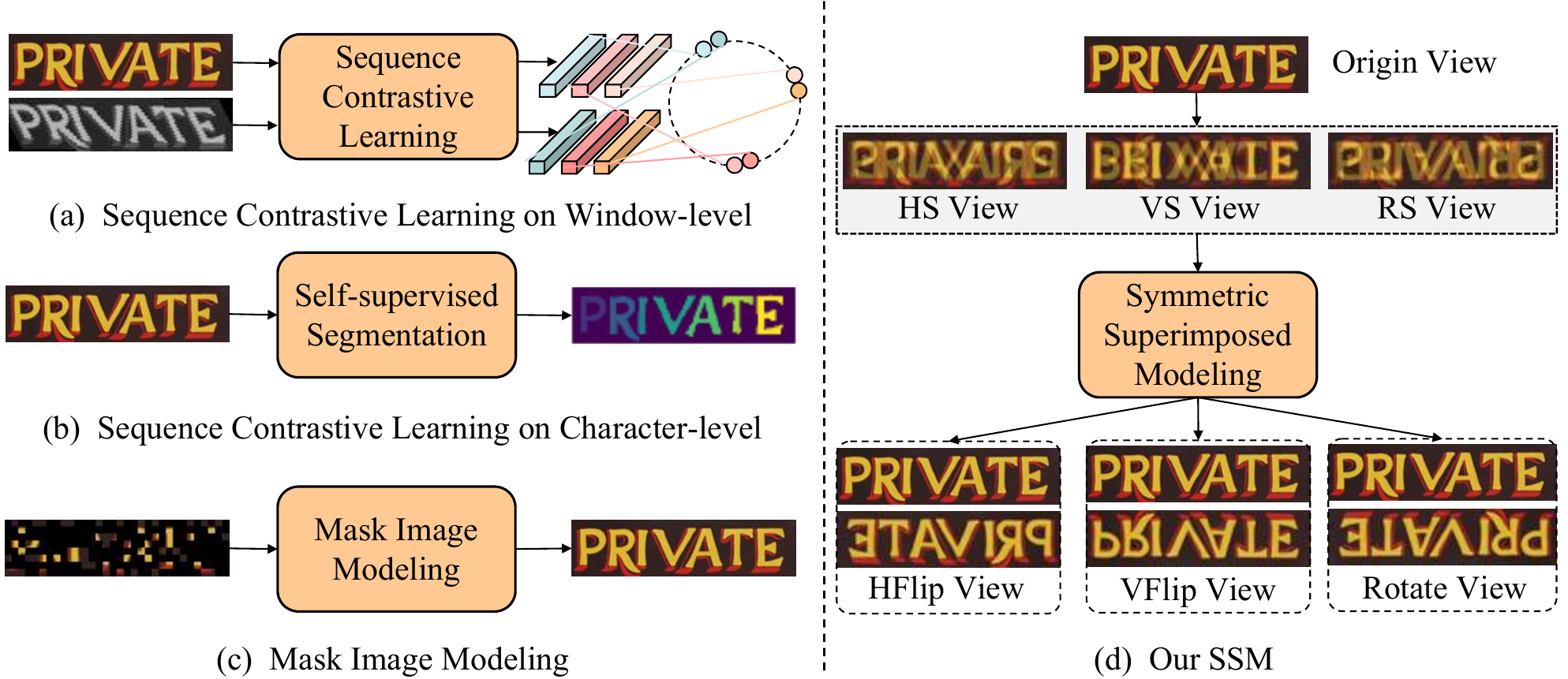} 
    \caption{The comparison with mainstream self-supervised text recognition methods and our SSM. Rotate, VFlip and HFlip Views stand for the symmetrically augmented image created through 180-degree rotation, vertical flipping, and horizontal flipping. HS, VS, and RS views respectively represent images formed by superimposing HFlip, VFlip, and Rotate View with the Origin View.}
    \label{fig: Introfig}
\vspace{-1.5em}
\end{figure}

\section{Introduction}\label{sec:Intro}
 Reading text from images is a fundamental and valuable task in computer vision with practical applications such as multi-modal analysis, visual search, self-driving cars, and more. Since labeled real text images are scarce and expensive,  various self-supervised text recognition methods have been utilized to exploit the intrinsic knowledge of unlabeled real data to alleviate the data scarcity issues. These self-supervised text recognition methods can be categorized into two main types: 1) Sequence Contrastive Learning (SCL), and 2) Mask Image Modeling (MIM). Benefiting from representation learning on unlabeled data, these methods have effectively enhanced Scene Text Recognition (STR) performance. However, these methods face the challenge of achieving linguistic learning, which is proven essential for text recognition (e.g. ABINet \cite{ShanchengFang2021ReadLH}
, LPV \cite{zhang2023linguistic}).

For SCL methods, SeqCLR \cite{AviadAberdam2021SequencetoSequenceCL} and CCD~\cite{Guan_2023_ICCV} are representative works. As shown in Fig.~\ref{fig: Introfig}(a), SeqCLR ensures the local representation consistency between the same instance window across the two augmented views. Fig.~\ref{fig: Introfig}(b) shows that CCD~\cite{Guan_2023_ICCV} further ensures the character-level representation consistency based on the self-supervised segmentation. Hence, both of the two methods essentially focus on performing discriminative consistency learning on local character representations.

For MIM-like methods,  MAERec~\cite{jiang2023revisiting} has attracted considerable attention in self-supervised text recognition. As discussed in MAERec, MIM-like methods essentially forces the model to infer the whole character from a few smallest parts of a character, due to covering a large portion of the text image. However, masking 75\% image patches drops nearly all text foreground areas, as shown in Fig.~\ref{fig: Introfig}(c). Thus, MIM-like methods also focus on local character features but overlook linguistic information (spelling rules between characters) in the text image.

Based on the analysis above, we can derive an observation that the previous self-supervised STR methods mainly focus on learning robust visual features of characters, overlooking the linguistic relationship between characters. Hence, it is meaningful to simultaneously capture character features and the implicit linguistic information in visual space.

 To this end, we propose a novel self-supervised learning paradigm, named \textbf{S}ymmetric \textbf{S}uperimposition \textbf{M}odeling (SSM). The pretext task of SSM is to reconstruct the direction-specific pixel and feature signals from the symmetrically superimposed input. We adopt a Siamese network with an online branch and target branch to implement SSM. Specifically, we first construct inverted images by randomly selecting from three inversion enhancement techniques: horizontal flip, vertical flip, and 180-degree rotation. Then we superimpose it onto the origin image to create the symmetric superimposed input. 
 For pixel reconstruction, we directly recover the original and inverted images (with online branch), as shown in Fig.~\ref{fig: Introfig}(d). The original and inverted images are the pixel targets to guide the decoupling of the superimposed input.  For feature reconstruction, we utilized the target branch to decouple the same symmetric superimposed input with irregular views, creating the original and inverted target feature on the fly. Subsequently, the online branch reconstruction the original and inverted target feature of the irregular view at the semantic level. We jointly use discriminative consistency loss and dense reconstruction loss to supervise the feature reconstruction process. Compared to MIM-like methods, we do not mask any patches and skillfully use symmetric superposition to disrupt character shapes and linguistic rules. Consequently, the pretext task of pixel and feature reconstruction from the superposed input can facilitate the learning of character shape features and linguistic information from the perspective of visual texture and feature semantics.
 In summary, our contributions mainly include:
 \begin{itemize}
    \item We propose a novel pre-training framework based on Symmetric Superimposition Modeling, which is the first self-supervised STR method dedicated to linguistic learning in visual space. 
     \item We present a dual architecture for the joint reconstruction at both the pixel and feature level. This design enables joint learning of character visual features and implicit linguistic information from the texture-level and semantic level, further improving representation quality.
    \item Experiments demonstrate that SSM achieves state-of-the-art performance on various text recognition benchmarks. The 4.1\% average performance gains on various STR methods also highlight the SSM's generality. Additionally, compared to other self-supervised methods, SSM has a 15.5\% and 1.5\% performance gain in multilingual text recognition with individual training and joint training settings, respectively.
\end{itemize}
\vspace{-.5em}

\section{Related Work}\label{sec:related work}
\subsection{Text Recognition}
Scene Text Recognition (STR) methods can be summarized into language-free and language-aware methods. Language-free methods~\cite{shi2016end,wang2017gated,ijcai2022p124} treat STR as a character classification task, focusing on how to extract robust visual features. Language-aware methods can leverage linguistic information to improve robustness. The attention-based methods~\cite{SRN,SCATTER:litman2020scatter,SATRN,NRTR} use various attention mechanisms to implicitly model linguistic rules. LISTER~\cite{LISTER} designs neighbor attention decoding for long text recognition. MGP-STR \cite{wang2022multi} and PARSeq \cite{bautista2022scene} focus on learning an internal Language Model(LM) during visual predicting. Some alternative approaches like ABINet~\cite{ShanchengFang2021ReadLH} and LevenOCR~\cite{da2022levenshtein} propose to refine the visual predictions via an external LM. 

\subsection{Self-Supervised Text Recognition}
Self-supervised learning technologies have been widely used for various works such as ML-LMCL \cite{cheng2023ml}, TKDF \cite{cheng2023accelerating} and MESM \cite{liu2024towards}. The most representative paradigms for computer vision are  Contrastive Learning~\cite{mocov3} and Masked Image Modeling (MIM)~\cite{he2022masked,bao2022beit,chen2023context}. Recently, these self-supervised methods have been adopted for text recognition. SeqCLR \cite{AviadAberdam2021SequencetoSequenceCL} is the pioneering work that proposes to model Contrast Learning on high-level sequence features for the first time. PerSec \cite{HaoLiu2022PerceivingSC} further performs Contrast Learning(CL) on both stroke-level features in shallow layers and Semantic-level features in deep layers. Later, DiG \cite{yang2022reading} proposes to learn discriminative and generative features by integrating the CL and MIM. To explicitly focus on character structures and ensure sequence consistency, Guan proposes a character-level self-distillation framework~\cite{Guan_2023_ICCV} based on unsupervised text segmentation maps. Recently, Union14M-U\cite{jiang2023revisiting}, a 10M scale unlabeled real data has been presented together with MAERec-S, which uses MAE for self-supervised text recognition. In addition, MaskOCR \cite{lyu2022maskocr} and DARLING \cite{zhang2024choose} use synthetic data for text recognition pre-training.
\vspace{-.5em}
\begin{figure*}[htb]
    \centering 
    \includegraphics[width=0.95\textwidth]{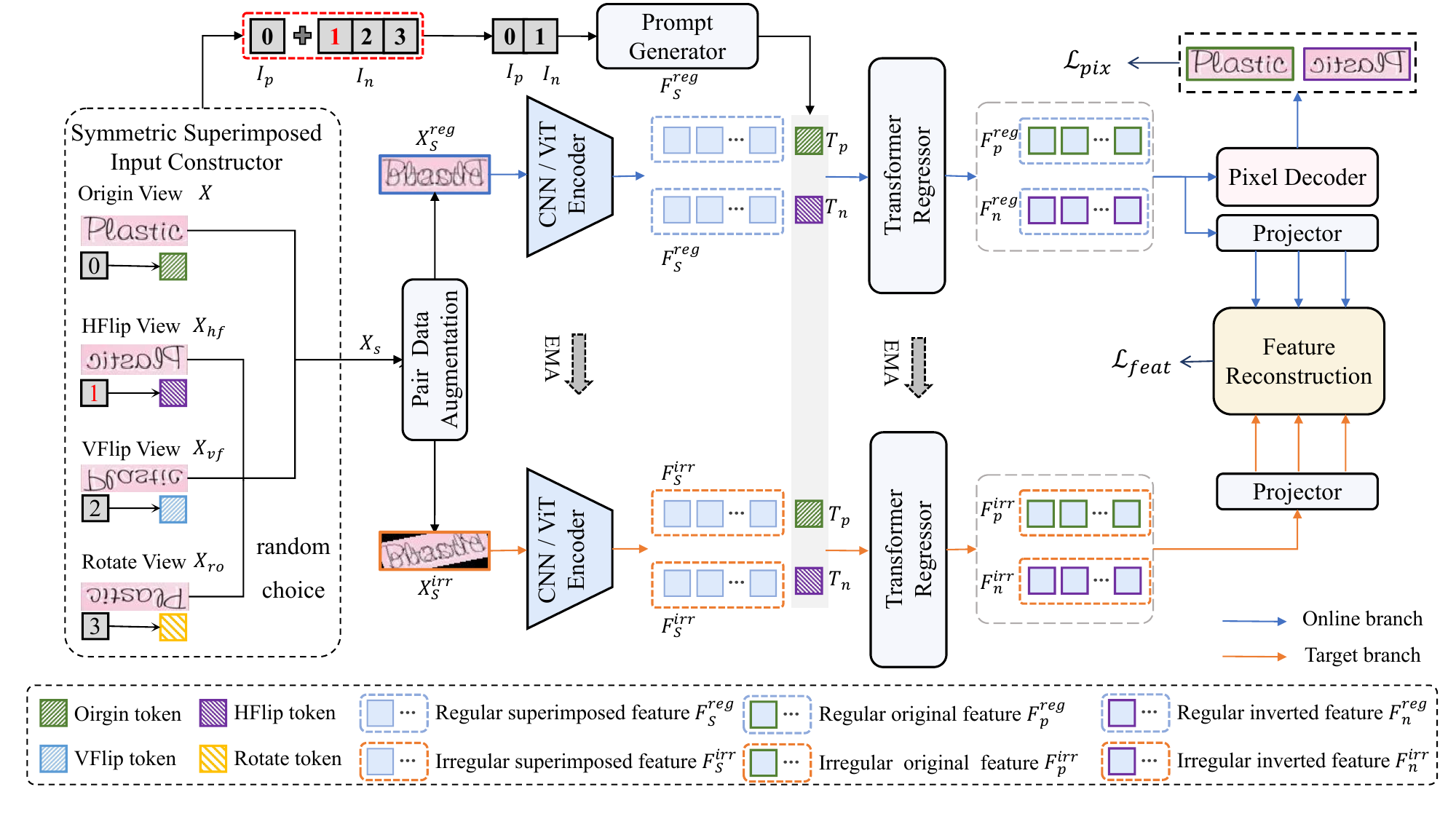} 
    \caption{The pre-training framework of SSM. The blue arrow and green arrow stand for the workflow of the online branch and target branch respectively. Origin View: original image, HFlip View: horizontally flipped image, VFlip View: vertically flipped image, Rotate View: 180-degree rotated image. $T_P$ and $T_n$ correspond to the original and the reversed text direction, respectively. }
    \label{fig: Pre-training Framework Overview}
\vspace{-0.5cm}
\end{figure*}

\section{Methodology}
The architecture of our proposed SSM is shown in Fig.~\ref{fig: Pre-training Framework Overview}, which comprises a Symmetric Superimposed Input Constructor, an online branch (blue arrow), and a target branch (orange arrow). We use the superimposed image of the original image and the horizontally flipped image as the input case to illustrate the whole workflow of our SSM.
\subsection{Symmetrically Superimposed Input}\label{subsec:architecture}
\noindent \textbf{Symmetrically Superimposed Input Constructor} 
To ensure a large character overlapping area of the superimposed inputs, we considered three symmetrical augmenting operations: Horizontal Flipping $\mathbf{Hf}(\cdot)$, Vertical Flipping $\mathbf{Vf}(\cdot)$, and 180-degree rotation $\mathbf{Ro}(\cdot)$. 
For the input image $X$, we randomly select one of  $\mathbf{Hf}(\cdot)$, $\mathbf{Vf}(\cdot)$, and $\mathbf{Ro}(\cdot)$ to obtain the corresponding inverted image $X_R$, where $X_R \in \{X_{hf}, X_{vf}, X_{ro}\}$. For the original image $X$, the original direction index $I_p$ is fixedly 
 set to 0. For the inverted image $X_R$, the inverted direction index $I_n$ is selected from $\{1, 2, 3\}$ according to the type of inverted image, where (1, 2, 3) is assigned to ($X_{hf}$, $X_{vf}$, $X_{ro}$), respectively. These indexes are subsequently encoded to guide direction-specific reconstruction. 
 Finally, we get the symmetrically superimposed input $X_S$ by superimposing the original image $X$ and its inverted images $X_{R}$. In the case of Fig.~\ref{fig: Pre-training Framework Overview}, $X_{hf}$ is selected and the $I_n$ is set to 1 (red color).

\noindent \textbf{Data Augmentation}
The superimposed image $X_{S}$ undergoes weak augmentations (\emph{e.g.}, gaussian blur, and grayscale conversion) to create a regular view $X^{reg}_{S}$ as the final input of online branch. The $X$ and $X_{R}$ are also transformed to $X^{reg}$ and $X^{reg}_{R}$ to supervise the image reconstruction of the online branch. For the target branch, we utilize the combination of both weak augmentations and geometry-based augmentations (\emph{e.g.}, affine transformation and perspective warping) to generate a pair of irregular views: $X^{irr}_{S}$, $X^{irr}$ and $X^{irr}_{R}$. 
\vspace{-5pt}
\subsection{Symmetric Superimposition Modeling }\label{subsec:mdr}
We use a Siamese Network with an online branch and a target branch to implement SSM for both pixel-level image reconstruction and feature-level representation reconstruction.

\noindent \textbf{1) Pixel-level Image Reconstruction}
We employ an Encoder-Regressor-Decoder architecture to perform the image reconstruction according to the specific direction index pair in the online branch. Specifically, we first leverage the ViT Encoder $\mathcal{F(\cdot)}$ to map the $X^{reg}_{S}$ as latent feature $F_{s}\in\mathbb{R}^{\frac{HW}{p^2} \times d}$, where the $p$ is the patch size and the $d$ is the embedding dim. Meanwhile, we utilize the Prompt Generator $\mathcal{G(\cdot)}$ to encode the direction index pair ($I_p$, $I_n$) into the direction prompt token pair ($T_p$, $T_n$) with the same dimension of the ViT Encoder.  
\begin{equation}
    \begin{aligned}
    T_p &= FFN(Embed((I_p)) \\
    T_n &= FFN(Embed((I_n)),
    \end{aligned}
\end{equation}
where the Prompt Generator $\mathcal{G(\cdot)}$ consists
 of an embedding layer, two-layer $FFN$ with normalization. 

After that, $T_p$ and $T_n$ are each concatenated with the latent feature of superimposed input $F_{s}$ and then sent into the Transformer Regressor $\mathcal{R(\cdot)}$ for feature decoupling in symmetric directions. The Transformer Regressor $\mathcal{R(\cdot)}$
 is a series of vision transformer blocks. Thanks to the global interaction capabilities of the attention mechanism, Regressor $\mathcal{R(\cdot)}$ can extract the original direction feature $F_p$ and the feature $F_n$ corresponding to symmetrically reversed direction from the mixed features $F_s$. The process can be formulated as follows:
 \begin{equation}
    \begin{aligned}
    F_p &= \mathcal{R}([T_p,  \mathcal{F}(X^{reg}_{S})])\in\mathbb{R}^{\frac{HW}{p^2} \times d} \\
    F_n &= \mathcal{R}([
    T_n,  \mathcal{F}(X^{reg}_{S})])\in\mathbb{R}^{\frac{HW}{p^2} \times d}
    \end{aligned}
\end{equation}

Next, we use a lightweight pixel decoder to predict the RGB pixels of the corresponding views $X^{reg}$ and $X^{reg}_{R}$ from latent features $F_p$ and $F_n$. To prevent the pixel decoder's parameters from dominating the learning process, we form the pixel decoder using only two linear layers with GELU. Finally, the $L_2$ loss function is adopted to optimize the image reconstruction process, and the complete loss function of pixel-level image reconstruction can be defined as follows:
\begin{equation}
L_{pix} = \frac{1}{N} \sum_{i=1}^{N} (\left\| y^i_p - \hat{y}^i_p \right\|^2 + \left\| y^i_n - \hat{y}^i_n \right\|^2)
\end{equation}
where $y^i_p, \hat{y}^i_p, \in R^{3}$ and $y^i_n, \hat{y}^i_n \in R^{3}$ are the prediction-target pair of $x^{reg}_p$ and $x^{reg}_n$, respectively; $N$ is the number of pixels.

\noindent \textbf{2) Feature-level Representation Reconstruction} To enhance the feature discriminability of character semantics and to model the spatial context between character visual semantics, we apply feature reconstruction in high-dimensional space.
Specifically, we add the projector $\mathcal{H(\cdot)}$ followed by the Regressor $\mathcal{R(\cdot)}$ in the online branch. $\mathcal{H(\cdot)}$  aggregates the decoupled features ($F_p$ and $F_n$) into window-level features by adaptive mean-pooling and then maps them into high-level feature space, forming the corresponding prediction feature queries ($Q_p$ and $Q_n$). To construct the target representation, we introduce a target branch with the same structure as the online branch, except for the pixel decoder. $X^{irr}_S$ is fed into the target branch to obtain the window-level target feature keys ($K_p$ and $K_n$) on the fly. 
Next, we jointly employ the discriminative consistency loss $L_{dis}$ and dense reconstruction loss $L_{den}$ to supervise the reconstruction from $Q_p$ to $K_p$ and $Q_n$ to $K_n$ across the regular and irregular views. The loss $L_{dis}$ is to enhance the  character classification representation at the semantic level, which can be formulated as:
\begin{small}
{\setlength\abovedisplayskip{1pt}
\setlength\belowdisplayskip{1pt}
\begin{equation}
\begin{aligned}
    L_{dis} =  &-log\frac{exp(Q_p \cdot K^{+}_p / \tau)}{exp(Q_p \cdot K^{+}_p / \tau) + \sum_{K^{-}_p}{exp(Q_p \cdot K^{-}_p / \tau)}} \\
           &-log\frac{exp(Q_n \cdot K^{+}_n / \tau)}{exp(Q_n \cdot K^{+}_n / \tau) + \sum_{K^{-}_n}{exp(Q_n \cdot K^{-}_n / \tau)}}
\end{aligned}
\end{equation}}
\end{small}
where $K^+$ stands for the matched positive samples and $K^-$ indicates the negative samples collected from the same batch for both ($Q_p$,$K_p$) and ($Q_n$,$K_n$). $\tau$ denotes the temperature. 

Since SiameseMIM~\cite{tao2023siamese} demonstrates that predicting dense representations helps improve sensitivity to global structures, we directly minimize the distance between the predicted and the target features by $L_{mse}$ loss, aiming to achieve semantic-level context modeling. Assuming that predictions are $Q=\{q^i \in \mathbb{R}^{d} | i=1, ..., \mathcal{N}\}$ and targets is $K=\{k^i \in \mathbb{R}^{d} | i=1, ..., \mathcal{N}\}$, where $\mathcal{N}$ is the total number of feature instances in one batch. The dense reconstruction loss $L_{den}$ ($L_{mse}$) between the original Q-K pairs ($Q_p$ and $K_p$) as well as inverted Q-K pairs ($Q_n$ and $K_n$) 
can be formulated as:
\begin{equation}
L_{den} = \frac{1}{\mathcal{N}} \sum_{i=1}^{\mathcal{N}} (\left\| q^i_p - k^i_p \right\|^2 + \left\| q^i_n - k^i_n \right\|^2)
\end{equation}

Finally, the optimization objectives of SSM are as follows:
\begin{equation}
    L =  L_{pix} + \alpha \times (\overbrace{L_{dis} + L_{den}}^{L_{feat}})
\end{equation}
where $\alpha$ is a scaling factor, the training objective of semantic-level feature reconstruction $L_{feat}$ is the sum of $L_{dis}$ and $L_{den}$.

\subsection{Downstream Tasks }\label{subsec:down}
For text recognition downstream tasks, we add a text decoder after the ViT Encoder inherited from SSM. The text decoder consists of 6 transformer blocks and a linear prediction layer with 96 channels to predict the final characters. Besides, the text super-resolution and text segmentation results are presented in Appendix \textred{B.1} and \textred{B.2}.

\begin{figure}
    \centering 
      \includegraphics[width=\linewidth]{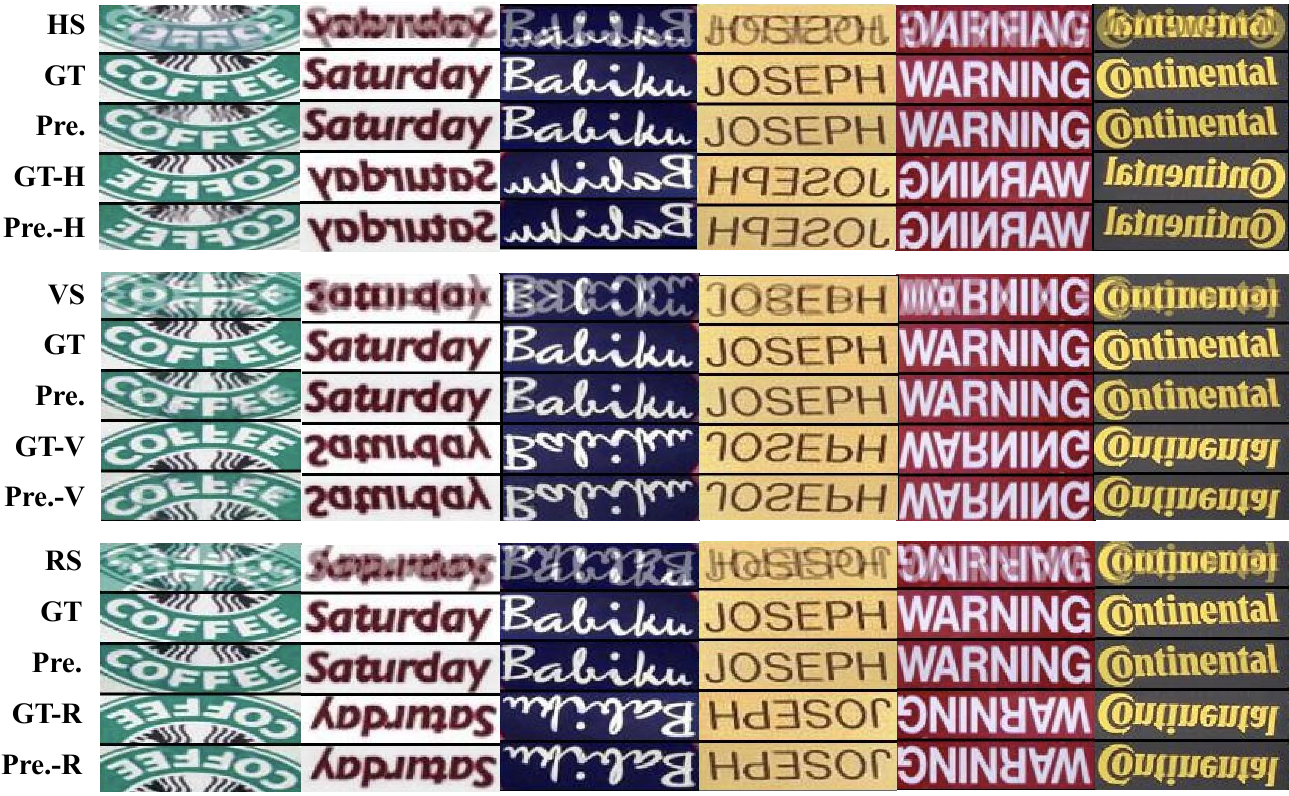} 
    \caption{Reconstruction Visualization. \textbf{GT}: the original image. \textbf{HS/ VS/ RS}: horizontal/ vertical / roteated superimposed input. \textbf{Pre.} indicates the pixel prediction of \textbf{GT}. \textbf{GT-H/ V/ R}: the inverted view of the \textbf{GT} (HFlip, VFlip, 180-degree rotation view, respectively). \textbf{Pre.-H/ V/ R}:the pixel prediction of \textbf{GT-H/ V/ R}.  }
    \label{fig: visualization}
\vspace{-1.5em}
\end{figure}
\begin{table*}[t]
\centering
\renewcommand{\arraystretch}{1.2}  
\footnotesize                      
\scalebox{0.93}{
\begin{tabular}{p{45mm}|c|p{12mm}p{12mm}p{12mm}p{12mm}p{12mm}p{12mm}|p{12mm}|p{10mm}}
\toprule
Method & Data & IIIT & SVT & IC13 & IC15 & SVTP & CUTE &Avg. &Params.\\
\hline
SeqCLR \textcolor{gray}{\cite{AviadAberdam2021SequencetoSequenceCL}} &STD & 82.9 &- &87.9 &- &- &- &- & -\\
SimAN \textcolor{gray}{\cite{SimAN}}&STD &87.5 &- &89.9 &- &- &- &- & -\\
PerSec-ViT \textcolor{gray}{\cite{HaoLiu2022PerceivingSC}}&STD &88.1 &86.8 &94.2 &73.6 &77.7 &72.7 & 83.8& - \\
\hline
DiG-ViT-Tiny \textcolor{gray}{\cite{yang2022reading}}        &STD & 95.8 &92.9 &96.4 &84.8 &87.4 &86.1 &91.8 & 20M\\ 
CCD-ViT-Tiny \textcolor{gray}{\cite{Guan_2023_ICCV}}         &STD &96.5  &93.4 &96.3 &85.2 &89.8 &89.2 &92.6 & 20M\\ 
\rowcolor{mygray}
SSM-ViT-Tiny  & STD  &96.5\wincrease{0.0} & 94.4\increase{1.0} & 96.3\wincrease{0.0} & 85.6\increase{0.4} & 89.3\decrease{0.5} & 89.9\increase{0.7} & 92.8\increase{0.2} & 20M\\                  
\hline
DiG-ViT-Small \textcolor{gray}{\cite{yang2022reading}}           &STD & 96.7 &93.4 &97.1 &87.1 &90.1 &88.5 &93.2 & 36M\\
CCD-ViT-Small \textcolor{gray}{\cite{Guan_2023_ICCV}}        &STD &96.8  &94.4 &96.6 &87.3 &91.3 &92.4 &93.6 & 36M\\
\rowcolor{mygray}
SSM-ViT-Small &STD & 97.4\increase{0.6} & 94.6\increase{0.2}  & 96.7\increase{0.1}  & 86.8\decrease{0.5}  & 91.3\wincrease{0.0}  & 94.8\increase{2.4}  & 93.8\increase{0.2} & 36M\\
\midrule
DiG-ViT-Tiny \textcolor{gray}{\cite{yang2022reading}}          &ARD & 96.4 &94.4 &96.2 &87.4 &90.2 &94.1 &93.4 & 20M\\ 
CCD-ViT-Tiny \textcolor{gray}{\cite{Guan_2023_ICCV}}         &ARD &97.1  &96.0 &97.5 &87.5 &91.6 &95.8 &94.2 & 20M\\
\rowcolor{mygray}
SSM-ViT-Tiny  & ARD  & 98.1\increase{1.0} & 96.1\increase{0.1} & 97.8\increase{0.3} & 89.0\increase{1.5} & 92.6\increase{1.0} & 96.5\increase{0.7} & 95.1\increase{0.9} & 20M\\
\hline
DiG-ViT-Small \textcolor{gray}{\cite{yang2022reading}}         &ARD &97.7 &96.1 &97.3 &88.6 &91.6 &96.2 &94.7 & 36M\\ 
CCD-ViT-Small \textcolor{gray}{\cite{Guan_2023_ICCV}}        &ARD &98.0  &96.4 &98.3 &90.3 &92.7 &98.3 &95.6 & 36M \\
\rowcolor{mygray}
SSM-ViT-Small          & ARD  & 98.9\increase{0.9} & 98.0\increase{1.6}  & 98.5\increase{0.2}  & 90.8\increase{0.5}  & 95.0\increase{2.3}  & 98.3\wincrease{0.0}  & 96.4\increase{0.8} & 36M\\
\bottomrule
\end{tabular}
}
\caption{Text recognition results compared to other self-supervised text recognizers. DiG, CCD, and SSM are all pre-trained on the OCR-CC.
}
\label{tab:self_supervised}
\vspace{-0.5em}
\end{table*}
\begin{table*}[t]
\centering
\renewcommand{\arraystretch}{1.2}
\setlength{\tabcolsep}{2mm}        
\resizebox{\linewidth}{!}{
\begin{tabular}{c|cccccccc|cccccc}
\toprule
\multirow{2}{*}{Method} & \multicolumn{8}{c|}{Union14M Benchmarks}                                                                                                                                        & \multicolumn{6}{c}{Other Challenge Datasets} \\ \cline{2-15} 
                        & Artistic & Contextless & Curve & General & \begin{tabular}[c]{@{}c@{}}Multi-\\ Oritented\end{tabular} & \begin{tabular}[c]{@{}c@{}}Multi-\\ Words\end{tabular} & Salient & Avg & TT    & CTW   & COCO  & ArT   & Uber  & Avg \\ \hline
Scratch-ViT-S       & 72.7     & 81.0        & 82.6  & 82.6    & 78.1                                                       & 81.7                                                   & 79.5    & 79.8 & 90.5  & 86.0  & 75.5  & 81.5  & 82.4  & 83.2 \\
MAE-ViT-S$^{\star}$ \textcolor{gray}{\cite{he2022masked}}          & 76.0     & 81.6        & 86.2  & 82.8    & 84.0                                                       & 83.5                                                   & 83.3    & 82.5 & 91.7  & 87.8  & 76.6  & 82.8  & 83.8  & 84.6 \\
DiG-ViT-S$^{\star}$ \textcolor{gray}{\cite{yang2022reading}}           & 77.4     & 82.5        & 85.9  & 83.8    & 83.5                                                       & 84.0                                                   & 84.3    & 83.0 & 91.7  & 87.2  & 77.7  & 83.4  & 84.9  & 85.0 \\
\rowcolor{mygray}
SSM-ViT-S           & \textbf{78.4}     & \textbf{84.7}        & \textbf{87.5}  & \textbf{84.0}    & \textbf{85.8}                                                       & \textbf{84.6}                                                   & \textbf{85.2}    & \textbf{84.3} & \textbf{92.9}  & \textbf{88.2}  & \textbf{78.1}  & \textbf{83.4}  & \textbf{86.5}  & \textbf{85.8} \\ \hline
$\uparrow$                         & \rbf{+5.7}     & \rbf{+3.7}        & \rbf{+4.9}  & \rbf{+1.4}    & \rbf{+7.7}                                                      & \rbf{+2.9}                                                   & \rbf{+5.7}    & \rbf{+4.5} & \rbf{+2.4}  & \rbf{+2.2}  & \rbf{+2.6}  & \rbf{+1.9}  & \rbf{+4.1}  & \rbf{+2.6} \\
$\triangle$                         & \gbf{+1.0}     & \gbf{+2.2}        & \gbf{+1.6}  & \gbf{+0.2}    & \gbf{+2.3}                                                       & \gbf{+0.6}                                                   & \gbf{+0.9}    & \gbf{+1.3} & \gbf{+1.2}  & \gbf{+1.0}  & \gbf{+0.4}  & \ybf{+0.0}  & \gbf{+0.6}  & \gbf{+0.8} \\ 
\bottomrule
\end{tabular}
}
\caption{Comparison with other self-supervised methods on Union14M benchmarks and other challenge datasets. $\uparrow$ and $\triangle$ represent the performance gains relative to the "train from scratch" model and the second-best model, respectively. $^{\star}$ stands for our implementation.}
\label{tab:union_ssr}
\vspace{-1.5em}
\end{table*}
\section{Experiments}
\subsection{Datasets}
\noindent \textbf{Unlabeled Pre-training Data} We utilize the latest unlabeled real-scene dataset Union14M-U for self-supervised learning, which contains 10 million instances collected from Book32, OCR-CC and OpenVINO. Besides, we also conduct pre-training on the complete OCR-CC dataset(15.77M unlabeled text images) to facilitate a fair comparison with works such as CCD ~\cite{Guan_2023_ICCV} and DiG~\cite{yang2022reading}.

\noindent\textbf{Text Recognition Fine-tuning Data} We use three types of labeled data. 1) STD: The synthetic data, comprising 14M images from MJSynth \cite{mj} and SynthText \cite{st}. 2) ARD: 2.78M annotated real data used by DiG and CCD. 3) Union14M-L: 3.6M real labeled data.

\noindent\textbf{Scene Text Recognition Benchmarks} The Common benchmarks include 
IC13, SVT, IIIT5K, IC15, SVTP and CUTE80. Considering the saturation issue with existing benchmarks, we leverage Union14M benchmarks~\shortcite{jiang2023revisiting} to evaluate. Some other challenging data such as TT, CTW, ArT, Uber, and COCO are also utilized to test.

\noindent \textbf{Multilingual Text Recognition Benchmarks.} Following MRN \cite{zheng2023mrn}, we tested the multilingual generalization capability of SSM on MLT19~\cite{nayef2019icdar2019}.

\subsection{Implementation Details}
\par{\noindent\textbf{Self-supervised Pre-training}} The pre-training is conducted on ViT, with image resolution of $32\times 128$, an AdamW optimizer, \textit{cosine} learning rate scheduler with a learning rate of 5e-4, batch size with 1,024, a weight decay of 0.05, $\beta_1=\text{0.9}$, $\beta_2=\text{0.95}$, and warm-up for 1 epoch in a total of 20 epochs. 
 
\par{\noindent\textbf{Text Recognition Fine-Tuning}} Our text recognition network is fine-tuned with STD or ARD or Union14M-L dataset. Patch size is $4 \times 4$. The text decoder consists of a 6-layer transformer block with an embedding dimension of 384. The batch size is 384 and the warm-up time is 1 epoch. The AdamW optimizer and a \textit{OneCycle} learning rate scheduler with a learning rate of 1e-4 are employed.

\noindent More datasets and implementation details are presented in Appendix \textred{A}.

\subsection{Comparasions with self-supervised methods.}
\noindent \textbf{Evaluation on Common benchmarks.} 
Tab.~\ref{tab:self_supervised} shows that SSM-ViT-Tiny outperforms the SeqCLR by 13.6\% and 9.4\% on IIIT and IC13, as well as outperforms PerSec-ViT (using 100M private data for pre-training) by 9\% on average accuracy. Besides
we compare with previous state-of-art self-supervised methods DiG and CCD, using the same
pre-training data, fine-tuning data (ARD), and network architectures (i.e., Tiny,
Small). Tab.~\ref{tab:self_supervised} illustrates that our methods achieve a new state-of-the-art average word accuracy on common benchmarks with 95.1\% and 96.4\% at the same model size. When fine-tuning with ARD, our methods achieve average performance gains of 0.9\% and 0.8\%. SSM-ViT-small even beats the CCD-ViT-Base which has a larger model size (96.4\% vs 96.3\%). The CCD attains high performance with explicit segmentation guidance, while our straightforward method surpasses it without needing extensive segmentation GT label preparation. These results demonstrate the effectiveness of guiding text image self-supervised pre-training from a linguistic learning perspective.

\noindent \textbf{Evaluation on Union14M benchmarks.} Considering the saturation of the common benchmarks, we further compare SSM with other self-supervised methods on more challenging benchmarks in Tab.~\ref{tab:union_ssr}. Although we established a strong baseline model Scratch-ViT-Small, and SSM-ViT-Small still achieved an average performance gain of 3.5\% and 2.6\% on Union14M and other challenging data, respectively. Additionally, SSM surpassed MAE and DiG by 1.8\% and 1.3\% of average word accuracy on the Union14M benchmarks. The performance gains on all of Union14M benchmarks' subsets demonstrate that SSM effectively learns intrinsic feature representations for different fonts, text orientations, and complex scenes. Surprisingly, SSM shows superior performance for contextless text images, which may be attributed to the fact that embedding linguistic learning in visual space can avoid the text distribution dependence of explicit text correction.
\begin{table}[h]
    \centering
    \begin{subtable}{\linewidth}
        \centering
\renewcommand{\arraystretch}{1.2}
\resizebox{\textwidth}{!}{
\begin{tabular}{c|cccccc|c}
\hline
Method & Arabic$^{\dag}$         & Korean         & Japanese      & Latin         & Chinese        & Bangla         & AVG            \\ \hline
Sctrach-S  & 2.3            & 7.2            & 11.2          & 79.4          & 1.6            & 6.1            & 17.2           \\
MAE-S   & 48.3           & 34.8           & 21.4          & 83.4          & 5.3            & 43.3           & 40.4           \\
DiG-S   & 48.6           & 37.9           & 23.5          & 83.6          & 4.7            & 50.3           & 41.4           \\
\rowcolor{mygray}
SSM-S   & \textbf{76.0}  & \textbf{60.5}  & \textbf{34.1} & \textbf{83.7}& \textbf{14.9}  & \textbf{72.6}  & \textbf{57.0}  \\
$\triangle$      & \textbf{\gbf{+27.4}} & \textbf{\gbf{+22.6}} & \textbf{\gbf{+10.6}} & \textbf{\gbf{+0.1}} & \textbf{\gbf{+10.2}} & \textbf{\gbf{+22.3}} & \textbf{\gbf{+15.5}} \\ \hline
\end{tabular}}
        \caption{Training each language data one by one.}
        \label{tab:mlt19_single}
    \end{subtable}
    \begin{subtable}{\linewidth}
        \centering
\renewcommand{\arraystretch}{1.2}
\resizebox{\textwidth}{!}{
\begin{tabular}{c|cccccc|c}
\hline
Method & Arabic$^{\dag}$         & Korean         & Japanese      & Latin         & Chinese        & Bangla         & AVG            \\ \hline
Sctrach-S  & 70.6            & 55.2            & 36.6          & 80.4          & 23.0            & 69.5            & 55.9           \\
MAE-S   & 72.1          & 63.5           & 42.7          & 82.9          & 30.1            & 72.8           & 60.7           \\
DiG-S   & 75.2           & 64.5           & 42.5          & 82.9          & 29.5            & 76.1           & 61.9           \\
\rowcolor{mygray}
SSM-S   & \textbf{77.2}  & \textbf{66.9}  & \textbf{44.2} & \textbf{83.7}& \textbf{31.7}  & \textbf{76.6}  & \textbf{63.4}  \\
$\triangle$      & \textbf{\gbf{+2.4}} & \textbf{\gbf{+1.5}} & \textbf{\gbf{+1.7}} & \textbf{\gbf{+0.8}} & \textbf{\gbf{+2.2}} & \textbf{\gbf{+0.5}} & \textbf{\gbf{+1.5}} \\ 
\hline
\end{tabular}}
        \caption{Joint training of all language data.}
        \label{tab:mlt19_joint}
    \end{subtable}
    \vspace{-1.8em}
    \caption{Comparison with other self-supervised methods on MLT19. $\dag$ stands for the reading of Arabic is naturally right-to-left.}
    \label{tab:mlt}
    \vspace{-12pt}
\end{table}

\noindent \textbf{Multilingual Text Recognition Performance.}
To further validate SSM's ability to learn linguistic information and its generality, we compare SSM with other self-supervised methods on MLT19 benchmarks. All methods are pre-trained on the Union14M-U for a fair comparison. In Tab.~\ref{tab:mlt19_single}, we finetune them on each language data one by one. Tab.~\ref{tab:mlt19_single} shows that our SSM outperforms the second-best method DiG by 15.5\% in the average accuracy (57.0\% v.s. 41.4\%). Besides, for the right-to-left Arabic language, SSM even outperforms the second-best method by 27.4\% (76.0\% v.s. 48.6\%), which is attributed to that SSM can capture linguistic information from right-to-left texts during the pre-training phase. These results demonstrate that SSM is an effective pretext task to capture linguistic information in multilingual texts and exhibits strong generalization and few-shot capabilities for data-scarcity languages. Tab.~\ref{tab:mlt19_joint} shows that all the methods benefit from jointly learning linguistic information of different languages, yet SSM still exhibits superior performance with 1.5\% average accuracy gains compared to DiG.

\subsection{Ablations and analysis}
\noindent \textbf{Effectiveness of architecture.} The third row of Tab.~\ref{tab:abl_arch} indicates that there is a 3.0\% average performance improvement on Union14M benchmarks with the pixel-level reconstruction, which need the combined effect of
prompt generator $\mathcal{G(\cdot)}$ transformer regressor $\mathcal{R(\cdot)}$ and pixel decoder $\mathcal{D(\cdot)}$. However, removing the regressor $\mathcal{R(\cdot)}$ results in a slight performance downgrade. The result demonstrates that the transformer regressor plays an indispensable role in SSM by guiding and disentangling the features of the original and inverted images.
Based on the pixel reconstruction, the Joint use of the EMA mechanism and project module $\mathcal{H(\cdot)}$ further brings accuracy gains of 1.3\% as it promotes learning discriminative character semantics and modeling spatial context at the semantic level. Here, $\mathcal{H(\cdot)}$ is mainly used to prevent excessive background noise in negative sample sampling and to map the decoupling feature to high-dimensional space. Lastly, the addition of Irregular view augmentations improves the average test accuracy up to 84.3\% due to diverse perspectives.

\noindent \textbf{Ablations on the types of
superimposed input.} We compared our superposition strategies with \textbf{\textit{add noise $\&$ blur}} and randomly \textbf{\textit{add another text image}}. As shown in Tab.~\ref{tab:abl_input}, the performance gains of \textbf{\textit{add another text image}} are close to that of \textbf{\textit{add noise $\&$ blur}}, both of which are lower than existing superposition strategies. We attribute this to the fact that the model can easily reconstruct the two images with inconsistent text content and different background styles from low-level pixel differences, limiting the exploration of intrinsic character semantics and linguistic information. We also found that individually superposing a 180-degree rotation view is more effective than individually superposing a specific flip view. The result suggests that the rotation influences both character orientation and sequence reading order, helping capture 
learning richer character semantics and spatial context relationships during reconstruction. Finally, the best performance (84.3\%) is achieved by combining HS, VS, and RS strategies, which improve the diversity of the superimposed input.
\begin{table}[t]
\centering
\renewcommand{\arraystretch}{1.0}
\resizebox{0.45\textwidth}{!}{
\begin{tabular}{cccccc|c}
\toprule
$\mathcal{G(\cdot)}$ & $\mathcal{R(\cdot)}$ & $\mathcal{D(\cdot)}$ & EMA & $\mathcal{H(\cdot)}$ &Irr view & Avg \\ \hline
\multicolumn{6}{l|}{\textit{Scractch training}}    &79.8 \\
\checkmark & & & & & &79.7 \\
\checkmark &\checkmark &\checkmark & & & &82.8 \\
\checkmark &\checkmark &\checkmark &\checkmark & & &83.6 \\
\checkmark &\checkmark &\checkmark &\checkmark &\checkmark & &84.1 \\
\rowcolor{mygray}
\checkmark &\checkmark &\checkmark &\checkmark &\checkmark &\checkmark&\textbf{84.3} \\
      
\bottomrule
\end{tabular}
}
\caption{Ablations about the effectiveness of architecture. $\mathcal{G(\cdot)}$, $\mathcal{R(\cdot)}$, $\mathcal{D(\cdot)}$ and $\mathcal{H(\cdot)}$ means prompt generator, regressor, pixel decoder, and projector, respectively. Testing on Union14M benchmarks.}
\label{tab:abl_arch}
\vspace{-1.em}
\end{table}
\begin{table}[htbp]
    \centering
    \begin{subtable}{0.48\linewidth}
        \centering
        \begin{tabular}{ccc|c}
        \toprule
        HS & VS & RS & Avg \\ \hline
        \multicolumn{3}{l|}{\textit{add noise $\&$ blur}}    &80.7     \\ 
        \multicolumn{3}{l|}{\textit{add another image}}    &80.4     \\ \hline
           \checkmark    &    &    &83.1     \\
               &\checkmark    &    &82.9     \\
               &    &\checkmark    &83.4     \\
           \rowcolor{mygray}
           \checkmark    &\checkmark    &\checkmark    &\textbf{84.3}    \\
        \bottomrule
        \end{tabular}
        \subcaption{Ablations on the types of superimposed input.}
        \label{tab:abl_input}
    \end{subtable}
    \hfill
    \begin{subtable}{0.46\linewidth}
        \centering
            \begin{tabular}{c|c}
            \toprule
            \begin{tabular}[c]{@{}c@{}} Reconstruction\\ Loss\end{tabular} & Avg   \\ \hline
            $L_{mse}$                                                        & 82.8 \\
            $L_{cos}$                                                         & 82.5 \\
            $L_{dino}$                                                  & 83.4 \\
            $L_{dis}$                                                  & 83.8 \\
            \rowcolor{mygray}
            $L_{dis}$+$L_{mse}$                                                   &\textbf{84.3} \\
            \bottomrule
            \end{tabular}
        \subcaption{Ablations on feature-level reconstruction loss. }
        \label{tab:abl_loss}
    \end{subtable}
    
    \vspace{-.5em}
    \caption{Ablation studies on the input type and consistency loss. HS, VS, and RS respectively means superimposing HFlip, VFlip, and Rotation View with the original image. }
    \label{tab:abl_input_loss}
    \vspace{-1.2em}
\end{table}

\begin{table*}[!h]
  \renewcommand{\arraystretch}{1.22}
  \resizebox{\linewidth}{!}{
    \begin{tabular}{cccccccccccccccccc}
      \hline
      \multirow{2}{*}{Type}                          & \multirow{2}{*}{Method} & \multicolumn{7}{c}{Common Benchmarks}                                                                                                                                                      & \multicolumn{9}{c}{Union14M-Benchmark}                                                                                                                                                                                                                                                             \\ \cline{3-18} 
                                                     &                         & \begin{tabular}[c]{@{}c@{}}IIIT\\ 3000\end{tabular} & \begin{tabular}[c]{@{}c@{}}IC13\\ 1015\end{tabular} & \begin{tabular}[c]{@{}c@{}}SVT\\ 647\end{tabular} & \begin{tabular}[c]{@{}c@{}}IC15\\ 2077\end{tabular} & \begin{tabular}[c]{@{}c@{}}SVTP\\ 645\end{tabular} & \begin{tabular}[c]{@{}c@{}}CUTE\\ 288\end{tabular}    & \multicolumn{1}{c|}{Avg}               & Cur                  & M-O  & Art        & Con         & Sal      & M-W                                       & Gen              & \multicolumn{1}{c|}{Avg}  & \multicolumn{1}{l}{Params.} \\ \hline
  \multirow{2}{*}{CTC}                           & CRNN \cite{shi2016end}                     & 90.8                                                & 91.8                                                & 83.8                                              & 71.8                                                & 70.4                                                   & 80.9                                     & \multicolumn{1}{c|}{81.6}              & 19.4                      & 4.5                                                     & 34.2            & 44.0                & 16.7                       & 35.7                                     & 60.4                 & \multicolumn{1}{c|}{30.7} & 8.3M                            \\
                                                  & SVTR ~\cite{ijcai2022p124}                   & 95.9                                                & 95.5                                                & 92.4                                              & 83.9                                                & 85.7                                                  & 93.1                                  & \multicolumn{1}{c|}{91.1}              & 72.4                       & 68.2                                                   & 54.1             & 68.0               & 71.4                       & 67.7                                     & 77.0                 & \multicolumn{1}{c|}{68.4} & 24.6M                            \\ \hline
  \multicolumn{1}{l}{\multirow{5}{*}{Attention}} &ASTER\cite{shi2018aster}                   & 94.3                                                & 92.6                                               & 88.9                                              & 77.7                                                & 80.5                                                   & 86.5                               & \multicolumn{1}{c|}{86.7}              & 38.4                      & 13.0                                                    & 41.8             & 52.9               & 31.9                       & 49.8                                    & 66.7                 & \multicolumn{1}{c|}{42.1} & -                            \\
  \multicolumn{1}{l}{}                           & NRTR\cite{NRTR}                  & 96.2                                                & 96.9                                                & 94.0                                              & 80.9                                                & 84.8                                                   & 92.0                                & \multicolumn{1}{c|}{90.8}              & 49.3                      & 40.6                                                    & 54.3              & 69.6               & 42.9                      & 75.5                                    & 75.2                 & \multicolumn{1}{c|}{58.2}                      & -                           \\

  \multicolumn{1}{l}{}                           & DAN\cite{DAN}                      & 95.5                                                & 95.2                                                & 88.6                                              & 78.3                                                & 79.9                                                   & 86.1                          & \multicolumn{1}{c|}{87.3}              & 46.0                       & 22.8                                                   & 49.3             & 61.6               & 44.6                       & 61.2                                   & 67.0                 & \multicolumn{1}{c|}{50.4} & -                               \\
  \multicolumn{1}{l}{}                           & SATRN\cite{SATRN}                  & 97.0                                                & 97.9                                                & 95.2                                              & 87.1                                                & 91.0                                                   & 96.2                             & \multicolumn{1}{c|}{93.9}              & 74.8                        & 64.7                                                  & 67.1             & 76.1               & 72.2                       & 74.1                                   & 75.8                 & \multicolumn{1}{c|}{72.1} & 55M                            \\
  \multicolumn{1}{l}{}                           & RobustScanner\cite{RobustScanner}            & 96.8                                                & 95.7                                                & 92.4                                              & 86.4                                                & 83.9                                                   & 93.8                          & \multicolumn{1}{c|}{91.2}              & 66.2                         & 54.2                                                  & 61.4            & 72.7                & 60.1                      & 74.2                                   & 75.7                 & \multicolumn{1}{c|}{66.4} & -                            \\ 
   \hline
  \multicolumn{1}{c}{\multirow{5}{*}{LM}}        & SRN\cite{SRN}                     & 95.5                                                & 94.7                                                & 89.5                                              & 79.1                                                & 83.9                                                   & 91.3                                & \multicolumn{1}{c|}{89.0}              & 49.7                        & 20.0                                                   & 50.7            & 61.0              & 43.9                        & 51.5                                  & 62.7                 & \multicolumn{1}{c|}{48.5} & 55M                            \\
  \multicolumn{1}{c}{}                           & ABINet\cite{ShanchengFang2021ReadLH}                  & 97.2                                                & 97.2                                                & 95.7                                              & 87.6                                                & 92.1                                                   & 94.4                                   & \multicolumn{1}{c|}{94.0}              & 75.0                          & 61.5                                                 & 65.3             & 71.1               & 72.9                      & 59.1                                 & 79.4                 & \multicolumn{1}{c|}{69.2} & 37M                            \\
  \multicolumn{1}{c}{}                           & VisonLAN\cite{YuxinWang2021FromTT}                & 96.3                                                & 95.1                                                & 91.3                                              & 83.6                                                & 85.4                                                   & 92.4                                   & \multicolumn{1}{c|}{91.3}              & 70.7                        & 57.2                                                   & 56.7             & 63.8               & 67.6                      & 47.3                                & 74.2                 & \multicolumn{1}{c|}{62.5} & 33M \\
  \multicolumn{1}{c}{}                           & MATRN\cite{MATRN}                   & 98.2                                                & 97.9                                                & 96.9                                              & 88.2                                                & 94.1                                                   & 97.9                                  & \multicolumn{1}{c|}{95.5}              & 80.5                        & 64.7                                                   & 71.1             & 74.8               & 79.4                      & 67.6                                 & 77.9                 & \multicolumn{1}{c|}{74.6} & 44M                           \\
  \multicolumn{1}{c}{}                           & PARSeq$^\star$\cite{bautista2022scene}                   & 98.0                                                & 96.8                                                & 95.2                                              & 85.2                                                & 90.5                                                   & 96.5                                  & \multicolumn{1}{c|}{93.8}              & 79.8                        & 79.2                                                   & 67.4             & 77.4               & 77.0                      & 76.9                                 & 80.6                 & \multicolumn{1}{c|}{76.9} & 24M                            \\
   \hline
  \multicolumn{1}{c}{\multirow{3}{*}{Pre-train}}      & MAERec-S\cite{jiang2023revisiting}                & 98.0                                                & 97.6                                                & 96.8                                              & 87.1                                                & 93.2                                                   & 97.9                                               & \multicolumn{1}{c|}{95.1}              & 81.4                         & 71.4                                                  & 72.0             & 82.0                & 78.5                     & 82.4                                 & 82.5                 & \multicolumn{1}{c|}{78.6} & 36M                            \\
  & DiG-ViT-Small$^\star$\cite{yang2022reading}                & 98.7                                                & 97.8                                                & 98.5                                              & 88.9                                                & 92.7                                                   & 96.5                                               & \multicolumn{1}{c|}{95.5}              & 85.9                         & 83.5                                                   & 77.4             & 82.5                & 84.3                     & 84.0                                 & 83.8                 & \multicolumn{1}{c|}{83.0} & 36M                            \\
  & MAERec-B\cite{jiang2023revisiting}                & 98.5                                                & 98.1                                                & \underline{97.8}                                              & \underline{89.5}                                                & \underline{94.4}                                                   & \underline{98.6}                                               & \multicolumn{1}{c|}{\underline{96.2}}              & \underline{88.8}                         & 83.9                                                  & \underline{80.0}             & \underline{85.5}                & 84.9                     & \underline{87.5}                                 & \textbf{85.8}                 & \multicolumn{1}{c|}{\underline{85.2}} & 142M                            \\
      \hline
  \multicolumn{1}{c}{\multirow{4}{*}{Ours}}      & SSM-ViT-Tiny                & 98.8                                                & 97.9                                                & 96.8                                              & 88.0                                                & 93.2                                                   & 97.2                                               & \multicolumn{1}{c|}{95.3}              & 81.7                         & 77.9                                                  & 72.3             & 79.7                & 79.7                     & 77.9                                 & 81.9                 & \multicolumn{1}{c|}{80.7} & 20M                            \\
    & SSM-ViT-Small                & \underline{99.0}                                                & 98.3                                                & \underline{97.8}                                              & \underline{89.5}                                                & 94.0                                                   & 98.3                                               & \multicolumn{1}{c|}{96.1}              & 87.5                         & \underline{85.8}                                                  & 78.4             & 84.8                & \underline{85.2}                     & 85.0                                 & 84.0                 & \multicolumn{1}{c|}{84.3} & 36M                            \\
         & SSM-ViT-Small-Turbo                & \textbf{99.1}                                                & \textbf{98.5}                                                & 97.7                                              & \textbf{89.9}                                                & \textbf{94.9}                                                   & \textbf{99.0}                                               & \multicolumn{1}{c|}{\textbf{96.5}}              & \textbf{91.0}                         & \textbf{89.4}                                                  & \underline{79.3}             & \textbf{86.0}                & \textbf{86.5}                     & \textbf{88.6}                                 & \underline{85.3}                 & \multicolumn{1}{c|}{\textbf{86.6}} & 36M                            \\
 \hline
      \end{tabular}}
  \caption{Performance of models fine-tuned on the \textbf{Union14M-L}. All pre-train and our methods are pre-trained on Union14M-U. \textbf{Bold} and underlined values stands for the 1st and 2nd results in each column. For a fair comparison, IC13 and IC15 are larger versions. Cur, M-O, Art, Ctl, Sal, M-W, and Gen respectively represent Curve, Multi-Oriented, Artistic, Contextless, Salient, Multi-Words, and General. Avg stands for the average word accuracy of corresponding benchmarks. $^\star$ represents our implementation. '*-Turbo' stands for training 20 epochs.}
  \label{tab:main_res}
  \vspace{-1.em}
\end{table*}


\noindent \textbf{Ablations on the feature-level reconstruction loss.} The first and second rows of 
 Tab.~\ref{tab:abl_loss} shows that only using distance measure functions such as \textit{MSE} $L_{mse}$ and \textit{Cosine} $L_{cos}$ results in lower word accuracy (82.5\% vs 94.3\%). We attribute this to the fact that only minimizing the feature distance between predictions and targets may lead to feature collapse. Besides, Tab.~\ref{tab:abl_loss} shows that our discriminative consistency loss $L_{dis}$ outperforms the dino loss $L_{dino}$ by 0.3\%. The combined use of $L_{dis}$ and $L_{mse}$ ultimately yielded the best results. $L_{dis}$ can mitigate the feature collapse of $L_{mse}$ by reducing the similarity between negative samples. Besides,  $L_{dis}$ focuses on enhancing discriminative features for character semantics while $L_{mse}$ is more sensitive to spatial structures, which helps model linguistic information in spatial context at the semantic level.
\vspace{-8pt}

\begin{figure}[htb]

\begin{minipage}[b]{.48\linewidth}
  \centering
\centerline{\includegraphics[width=4.4cm]{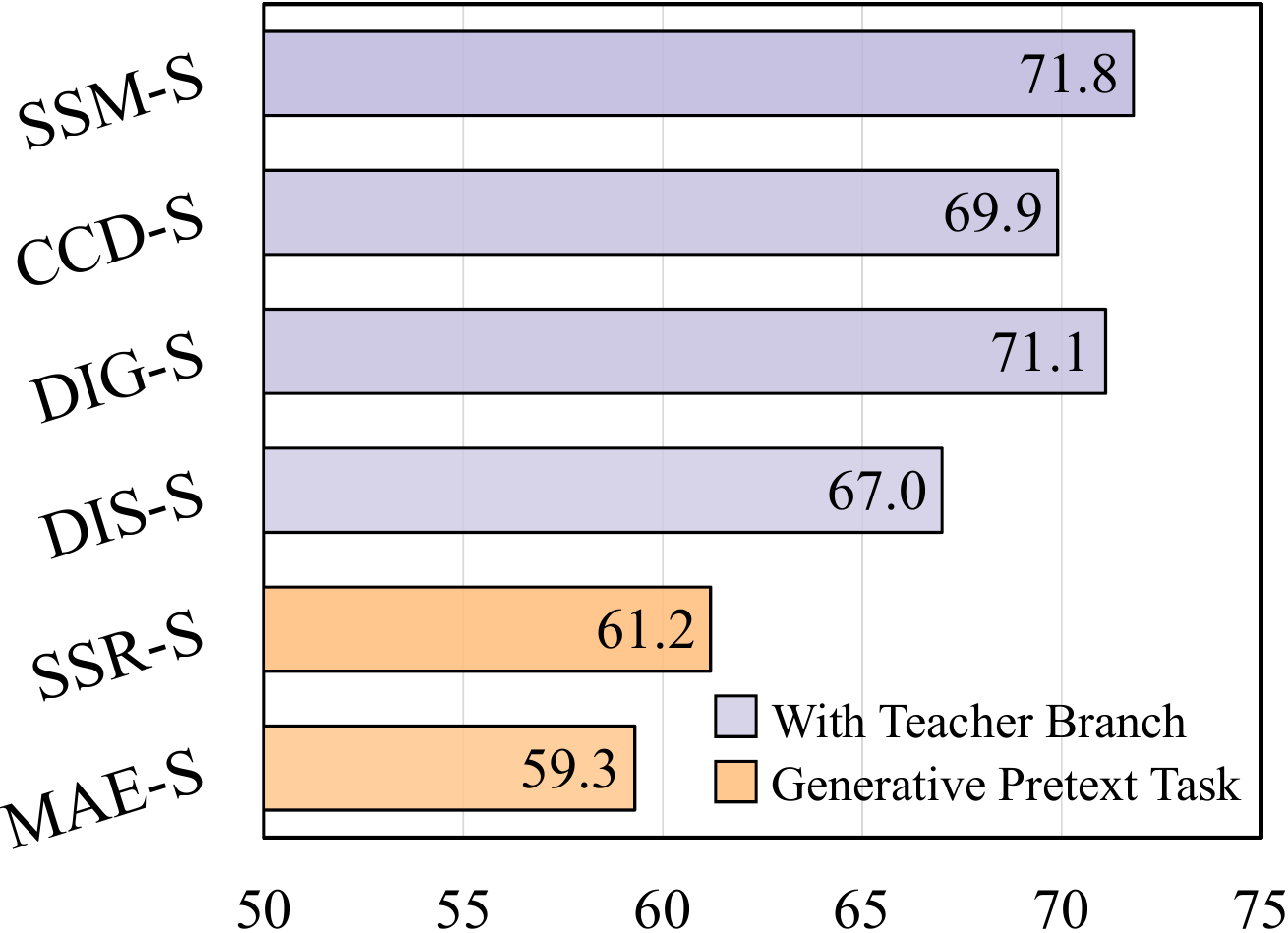}}
  \caption{Comparison of feature representation evaluation.}\medskip
  \label{fig:feat_rep}
\end{minipage}
\hfill
\begin{minipage}[b]{0.48\linewidth}
  \centering
\centerline{\includegraphics[width=4.2cm]{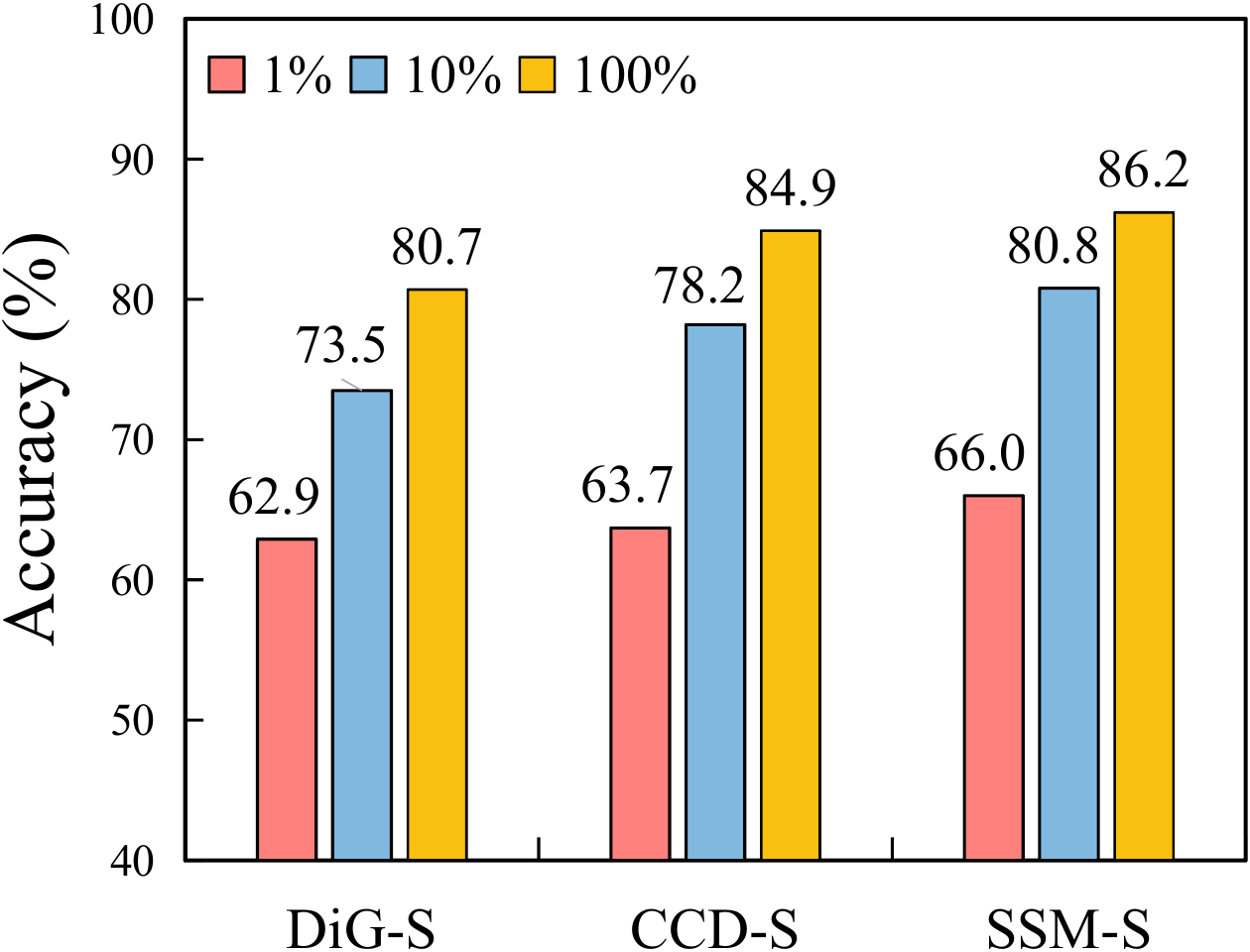}}
  \caption{Fine-tuning on ARD with different ratios.}\medskip
  \label{fig:data_ratio}
\end{minipage}
%
\vspace{-1.5em}
\end{figure}
\noindent \textbf{Feature representation evaluation.} We freeze the encoder and train the decoder with ARD, following the test setting of DiG. Fig.~\ref{fig:feat_rep} shows that the pixel-level reconstruction of SSM (SSR-S) has better feature representation than MAE-S (61.2\% vs59.3\%). By introducing feature-level representation reconstruction, SSM outperforms DiG by 0.7\%. We attribute this to strengthening the implicit linguistic representation through texture-level and semantic-level reconstruction.

\noindent \textbf{Fine-tuning with different data ratios.} We further fine-tune our method with 1\%(27.8K), 10\%(278K), and 100\%(2.78M) of ARD for a fair comparasion. Fig.~\ref{fig:data_ratio} shows that SSM-ViT-Small
outperforms the previous SOTA methods by 2.3\%, 2.6\% and 1.3\%, respectively. 

\begin{figure}[t]
\centering
\includegraphics[width=0.45\textwidth]{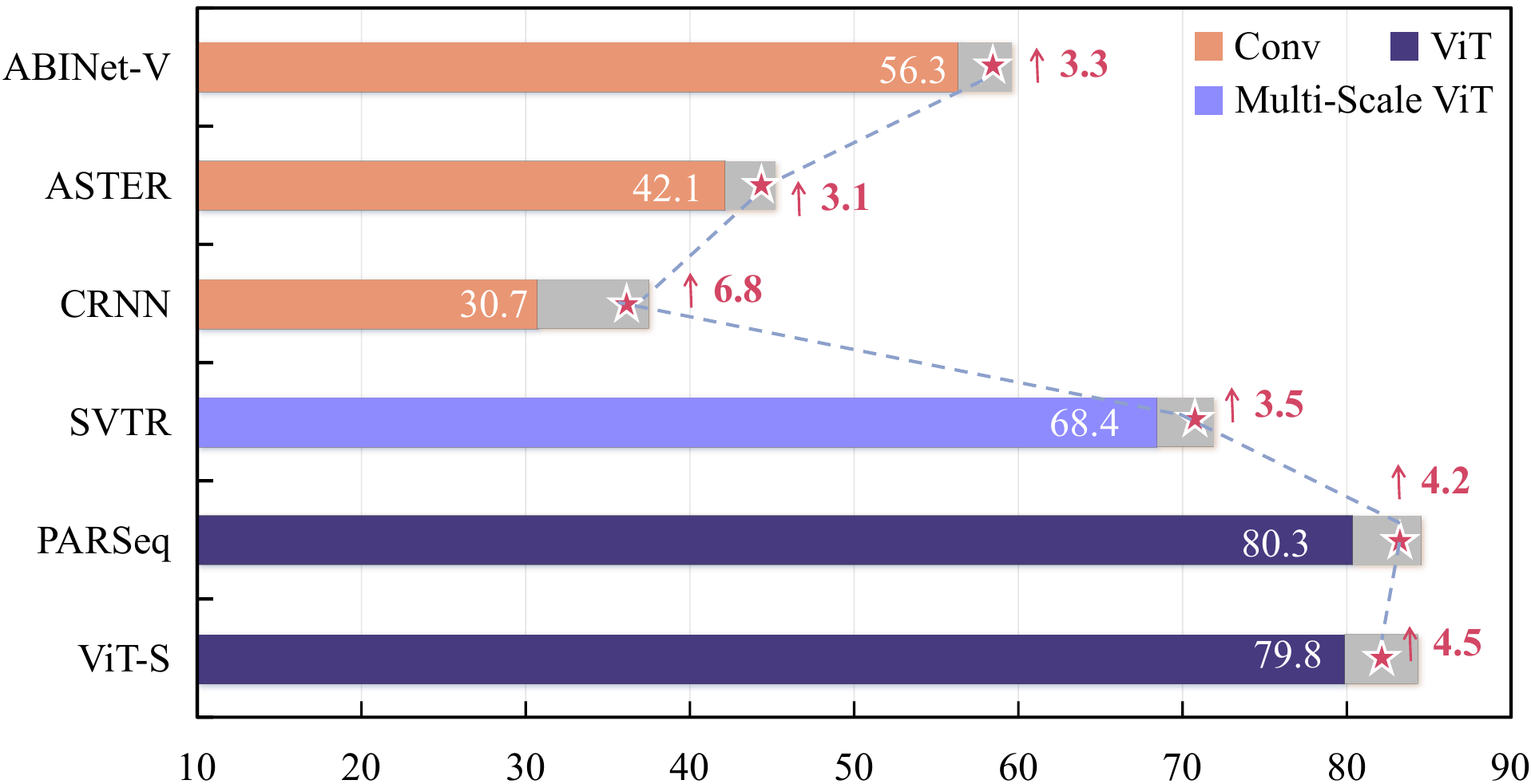}
\caption{Performance gains of SSM for various STR Methods. }
\label{fig:improve_str}
\vspace{-10pt}
\end{figure}

\noindent \textbf{Improvement of Various STR methods.}
Benefiting from masking-free operations, our method is not troubled by information leakage during feature downsampling. Consequently, it can be flexibly applied to various types of STR Methods, including Conv-based methods (e.g., ABINet-V, ASTER, and CRNN), multi-scale attention methods (e.g., SVTR), and vanilla ViT methods (e.g., our baseline ViT-*, and PARSeq). Fig.~\ref{fig:improve_str} shows that SSM can bring a performance gain of 4.1\% to the above STR method on Union14M benchmarks.

\begin{figure}[t]
\centering
\includegraphics[width=0.44\textwidth]{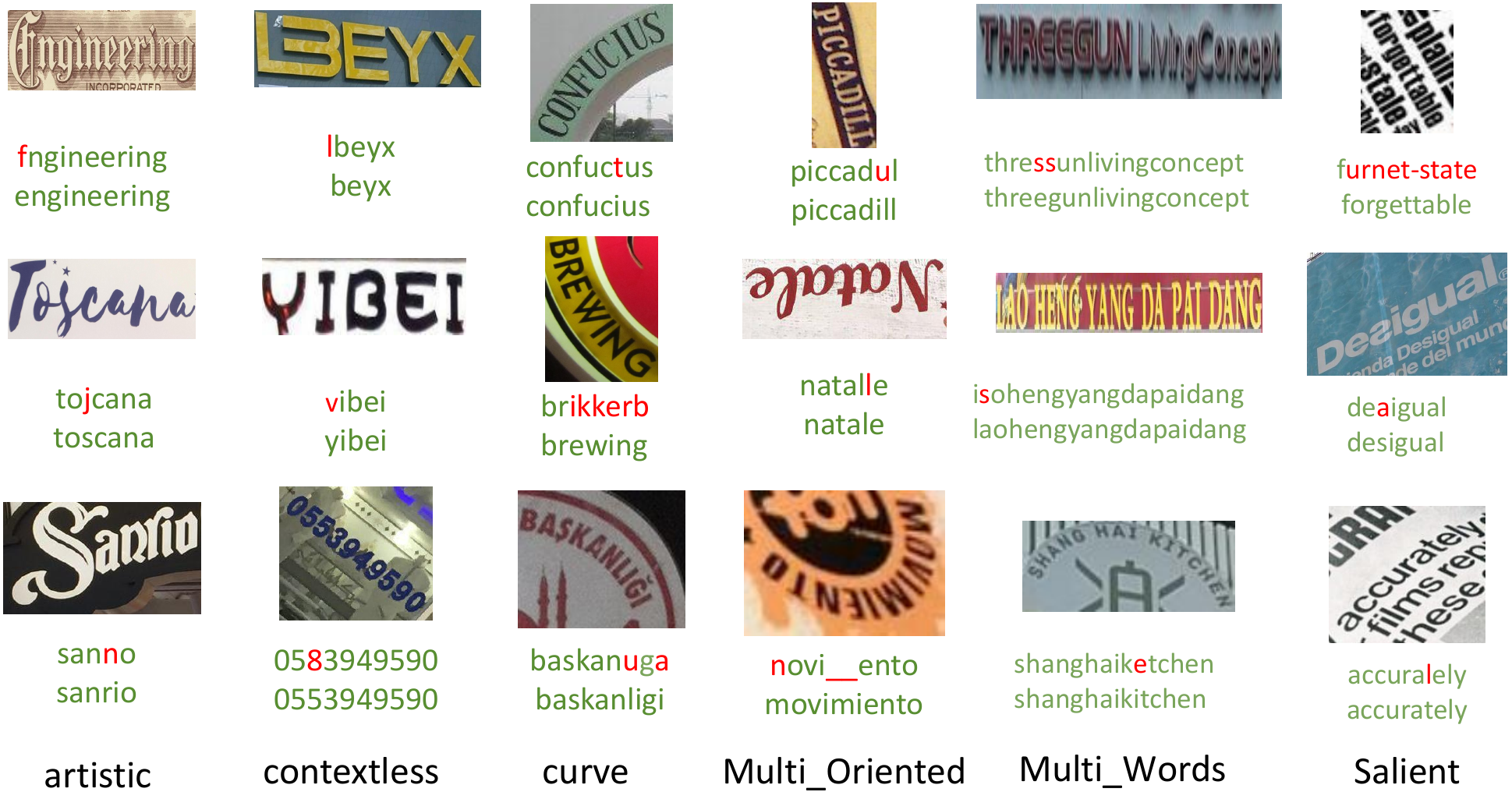}
\caption{Qualitative recognition results. The
top and bottom strings are predicted by DiG-S and SSM-S, with red indicating errors. }
\label{fig:rec_vis}
\end{figure}

\noindent \textbf{Qualitative recognition results.} Fig.~\ref{fig:rec_vis} indicates that SSM is robust to reversed text and complex-textured images in comparison. It also can infer complex characters based on contextual linguistic information.
\subsection{Comparisons with State-of-the-Arts}
In Tab.~\ref{tab:main_res}, we present a comparison between SSM and previous SOTA text recognition methods. Specifically, SSM-ViT-Tiny outperforms all supervised state-of-the-art (SOTA) methods (80.7\% vs 76.9\%) on more challenging Union14M benchmarks with only 20M parameters. Moreover, SSM-ViT-Small outperforms the other self-supervised pre-train methods with the same model size by 0.6\% and 1.3\% average performance. Surprisingly, by training 20 epochs, SSM-ViT-Small-Turbo pushes the new SOTA average performance on Common benchmarks and Union14M benchmarks to 96.5\% and 86.6\%, respectively. Note that SSM-ViT-Small-Turbo outperforms MAERec-B by 0.3\% and 1.4\% with only one-fourth of the parameters of the MAERec-B (36M vs 142M).
\vspace{-12pt}
\section{Conclusion}
In this paper, we propose a novel self-supervised text
recognition framework, termed SSM. In contrast to previous self-supervised text recognition methods that only focus on local visual features, SSM models the global spatial context for implicit linguistic learning. The objective of SSM is to reconstruct direction-specific pixel and feature signals from symmetrically superimposed text images. In this way, SSM captures both the local character feature and implicit linguistic information. Eventually, SSM improves the text recognition performance of various Scene Text Recognition methods and refreshes state-of-the-art performance on various text recognition benchmarks. 
\section{Acknowledgments}
This work is supported by the National Key Research and Development Program of China (2022YFB3104700), the National Nature Science Foundation of China (62121002,
U23B2028, 62102384). This research is supported by the Supercomputing Center of the USTC. We also acknowledge the GPU resource support offered by the MCC Lab of Information Science and Technology Institution, USTC.
\appendix

\bibliographystyle{named}
\bibliography{ijcai24}

\newpage
\appendix
\section*{Appendix}
\section{Datasets and Implementation Details}
\subsection{Datasets}

\noindent \textbf{Multilingual Text Recognition Benchmarks.} Following MRN \cite{zheng2023mrn}, we tested the multilingual generalization capability of SSM on MLT19~\cite{nayef2019icdar2019}.

\noindent \textbf{Text Segmentation Benchmark} Following DiG and CCD, we utilize TextSeg~\cite{textseg} as text segmentation benchmark.  TextSeg is a fine-annotated text dataset for text segmentation. It contains 4024 images (training set: 2646, validation set: 340, testing set: 1038). 

\noindent \textbf{Text Image Super-Resolution Benchmark} Following DiG and CCD, we evaluate the text image super-resolution ability of SSM on TextZoom dataset~\cite{textzoom}. TextZoom is a paired scene text super-resolution dataset. The training set contains 17367 pairs of images. The test dataset is split into three parts (easy, medium, and hard) according to the difficulty, which is composed of 1619, 1411, and 1343 pairs of images, respectively. 

 Besides, Tab.~\ref{appendixtab:datasets}presents detailed information about the datasets used for pretraining, text recognition fine-tuning, and performance validation in our study.

\subsection{Implementation Details}
In this section, we provide the implementation details of the self-supervised pre-training stage and supervised downstream tasks including Text Recognition, Text Super-resolution, and Text Segmentation. Tab.~\ref{appendixtab:downstream_arch} shows the decoder configurations of different text-related downstream tasks.
\par{\noindent\textbf{Self-supervised Pre-training.}} The pre-training is conducted on vanilla ViT, with image resolution of $32\times 128$, an AdamW optimizer, \textit{cosine} learning rate scheduler with a base learning rate of 5e-4, batch size with 1,024, a weight decay of 0.05, $\beta_1=\text{0.9}$, $\beta_2=\text{0.95}$, and warm-up for 1 epoch in a total of 20 epochs. All experiments are conducted with 8 NVIDIA 4090. Due to constraints in computational resources and training costs, we only adopt two variants: ViT-Tiny and ViT-Small. In addition, to validate the generality of SSM, we conducted additional pretraining and fine-tuning experiments (as shown in Fig.6 of main paper) on the backbones of CRNN~\cite{shi2016end}, ASTER~\cite{shi2018aster}, ABINet-Vision~\cite{ShanchengFang2021ReadLH}, and SVTR~\cite{ijcai2022p124}.

\par{\noindent\textbf{Text Super-resolution Fine-Tuning.}}
 We first rescale the input images to $32 \times 128$ with BICUBIC sampling, and then feed them to our super-resolution model (ViT Encoder with 3-layer attention block as decoder). The batch size is 96, the total number of fine-tuning epochs is 100, and the warm-up time is 10 epochs. The AdamW optimizer and a \textit{Cosine} learning rate scheduler with a learning rate of 1e-4 are employed. The general Peak Signal-to-Noise Ratio (PSNR) and Similarity Index Measure (SSIM) are employed to evaluate the quality of super-resolution images.
 
\par{\noindent\textbf{Text Segmentation Fine-Tuning.}}
Considering that TextSeg~\cite{textseg} is proposed for text segmentation on the whole image, while our SSM is performed on the cropped text region. Therefore, we crop text regions based on the bounding box annotations of text instances. In both training and evaluation periods, the text patches are resized to $32 \times 128$.
The total training epochs are 100 epochs and the warm-up time is 10 epochs following CCD. The batch size is set to 96. IoU Metric is utilized to evaluate the performance.
\begin{table}[t]
\centering
\footnotesize
\renewcommand{\arraystretch}{1.0}
\resizebox{0.48\textwidth}{!}{
\begin{tabular}{c|c|c|c}
\hline
Group                                     & Dataset            & stage                         & number        \\ \hline
OCR-CC                                    & OCR-CC             & pre-training                  & 15,759,229    \\ \hline
\multirow{3}{*}{Union14M-U}               & OCR-CC(filtered)   & \multirow{3}{*}{pre-training} & 5,596,389     \\
                                          & OpenVINO(filtered) &                               & 2,367,418     \\
                                          & Book32(filtered)   &                               & 2,718,509     \\ \hline
\multirow{2}{*}{STD}                      & MJSynth            & \multirow{2}{*}{fine-tuning}  & 8,919,241     \\
                                          & SynthText          &                               & 6,975,301     \\ \hline
\multirow{2}{*}{ARD}                      & TextOCR            & \multirow{2}{*}{fine-tuning}  & 708,264       \\
                                          & OpenVINO           &                               & 2,071,907     \\ \hline
Union14M-L                                & Union14M-L         & fine-tuning                   & 3,230,742     \\ \hline
\multirow{6}{*}{Common Benchmarks}        & IIIT5k             & \multirow{6}{*}{eval}         & 3,000         \\
                                          & SVT                &                               & 647           \\
                                          & IC13               &                               & 857 / 1,015   \\
                                          & IC15               &                               & 1,811 / 2,077 \\
                                          & SVTP               &                               & 645           \\
                                          & CUTE               &                               & 288           \\ \hline
\multirow{7}{*}{Union14M Benchmarks}      & Artistic           & \multirow{7}{*}{eval}         & 900           \\
                                          & Contextless        &                               & 779           \\
                                          & Curve              &                               & 2426          \\
                                          & General            &                               & 399274        \\
                                          & Multi-Oritented    &                               & 1369          \\
                                          & Multi-Words        &                               & 829           \\
                                          & Salient            &                               & 1585          \\ \hline
\multirow{5}{*}{Other Challenge Datasets} & COCO               & \multirow{5}{*}{eval}         & 9888          \\
                                          & ArT                &                               & 35,149        \\
                                          & Uber               &                               & 80,418        \\
                                          & Totaltext(TT)      &                               & 2,199         \\
                                          & CTW                &                               & 1,570         \\ \hline
\multirow{2}{*}{OST}                      & HOST               & \multirow{2}{*}{eval}         & 2,416         \\
                                          & WOST               &                               & 2,416         \\ \hline
\end{tabular}
}
\caption{The datasets used in the pre-training stage, fine-tuning stage, and evaluation stage of scene text recognition tasks.}
\label{appendixtab:datasets}
\vspace{-1.em}
\end{table}
\begin{table}[ht]
\centering
\renewcommand{\arraystretch}{1.0}
\resizebox{0.45\textwidth}{!}{
\begin{tabular}{c|clclcl}
\toprule
\multirow{2}{*}{Fine-tuning Stage} & \multicolumn{6}{c}{Decoder Configuration}                                              \\  \cline{2-7}
                            & \multicolumn{2}{c}{Embed\_dim} & \multicolumn{2}{c}{Depth} & \multicolumn{2}{c}{Heads} \\ \hline
Text Recognition             & \multicolumn{2}{c}{384}        & \multicolumn{2}{c}{6}     & \multicolumn{2}{c}{8}     \\ \hline
Text Segmentation            & \multicolumn{2}{c}{384}        & \multicolumn{2}{c}{3}     & \multicolumn{2}{c}{2}     \\ \hline
Text Image Super-Resolution  & \multicolumn{2}{c}{384}        & \multicolumn{2}{c}{3}     & \multicolumn{2}{c}{2}     \\ \bottomrule
\end{tabular}
}
\caption{The decoder configurations of different downstream tasks.}
\label{appendixtab:downstream_arch}
\vspace{-1.em}
\end{table}
\section{More Experiment Analysis
}
\subsection{Text Segmentation}
Following DiG, we perform text region cropping on the TextSeg~\cite{textseg} dataset and then proceed with fine-tuning and validation. We separately test the models pre-trained on OCR-CC and Union14M-U. In Tab.~\ref{appendixtab:text_segmentation}, all methods are trained for 100 epochs with a warm-up period of 10 epochs, without any data augmentations. In our reproduced results, Scratch-ViT-Small and the results reported by DiG are close, while there are 2-3\% performance gaps compared to the reported results of  DiG~\cite{yang2022reading} and CCD~\cite{Guan_2023_ICCV}.  It's worth noting that the pre-trained models on OCR-CC used in our replication are obtained from the open-source repositories of DiG and CCD. Considering that DiG and CCD do not open-source their code and fine-tuning weights for text segmentation, we speculate that this performance gap may be due to more data augmentation and training epochs used in DiG and CCD. Although there is a gap between the replicated results and the reported results, comparing results are still fair and convincing as they are based on the same codebase and training settings. When pretrained on OCR-CC, SSM brings a 4.5\% performance gain compared to baseline model Scratch-ViT-Smal$^\star$, outperforming the second-best model CCD-ViT-Small$^\star$ by 0.9\%.

Fig.~\ref{appendixfig:vis_seg} shows some
visualized text segmentation results. We can oberserve that our methods SSM performs better segmentation
effects than “Scratch"  methods on complex occasions where text
images suffer from background noise, perspective distortion.
\begin{table}[ht]
\centering
\renewcommand{\arraystretch}{1.0}
\resizebox{0.42\textwidth}{!}{
\begin{tabular}{c|c}
\toprule
{Method} & IoU(\%) \\ \hline
Scratch-ViT-Small       & 78.1                               \\
DiG-ViT-Small\textcolor{gray}{\cite{yang2022reading}}          & 83.1                                \\
CCD-ViT-Small\textcolor{gray}{\cite{Guan_2023_ICCV}}           & \textbf{84.8}                               \\
\hline
Scratch-ViT-Small$^\star$       & 78.4                          \\
DiG-ViT-Small$^\star$\textcolor{gray}{\cite{yang2022reading}}           & 81.3                           \\
CCD-ViT-Small$^\star$\textcolor{gray}{\cite{Guan_2023_ICCV}}           & 82.0                           \\
\rowcolor{mygray}
SSM-ViT-Small           & \textbf{82.9}                         \\ \hline
$\uparrow$  &\rbf{+4.5}  \\
\bottomrule
\end{tabular}
}
\caption{Comparison results on text segmentation experiments. The evaluation metric is the Intersection over
Union (IoU). ’$^\star$‘ stands for our implementation based on released pre-trained weights }
\label{appendixtab:text_segmentation}
\vspace{-1.em}
\end{table}
\subsection{Text Super-resolution}
Similar to the text segmentation experiments, DiG and CCD did not open-source their code and fine-tuning weights for text super-resolution. Hence, we reproduce the text super-resolution fine-tuning based on the same codebase, and the pre-trained models are obtained from the corresponding open-source repositories. For the PSNR metric, our reproductions of DiG and CCD yield consistent or even higher results compared to the reported ones. However, there is a gap between our reproduced results and the reported ones for the SSIM metric. We attribute this to the more data augmentations and training epochs used in reported results. 

Whatever, comparing results based on same reproduced codebase and training settings provides a fairer assessment or have artistic fonts. Tab.~\ref{appdixtab:sr_ocrcc} presents the comparative results when all methods are pre-trained on OCR-CC. For the SSIM metric, SSM-S outperforms the reproduced DiG-S and CCD-S (0.7646 vs. 0.7424 and 0.7646 vs. 0.7407), as well as the reported DiG-S (0.7646 vs. 0.7522). For the PSNR metric, SSM-S surpasses the second-best reproduced model DiG (22.19 vs. 21.88) and the best reported model CCD (22.19 vs. 21.84).  Notably, our SSM-S also  outperforms the previous state-of-the-
art text super-resolution expert model (22.19 vs. 21.49),  despite employing just three transformer units as decoder without additional designs. 

\begin{figure}[ht]
\centering
\captionsetup[subfigure]{justification=centering}
\begin{subfigure}[b]{0.48\textwidth}
         \centering
         \includegraphics[width=\textwidth]{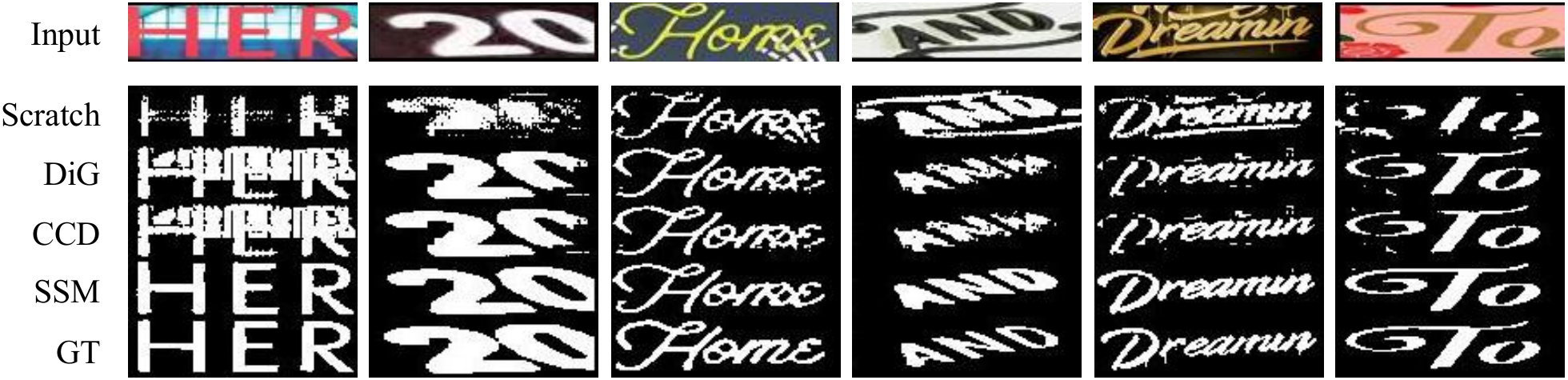}
         \caption{Text segmentation results}
         \label{appendixfig:vis_seg}
\end{subfigure}
\begin{subfigure}[b]{0.48\textwidth}
         \centering
         \includegraphics[width=\textwidth]{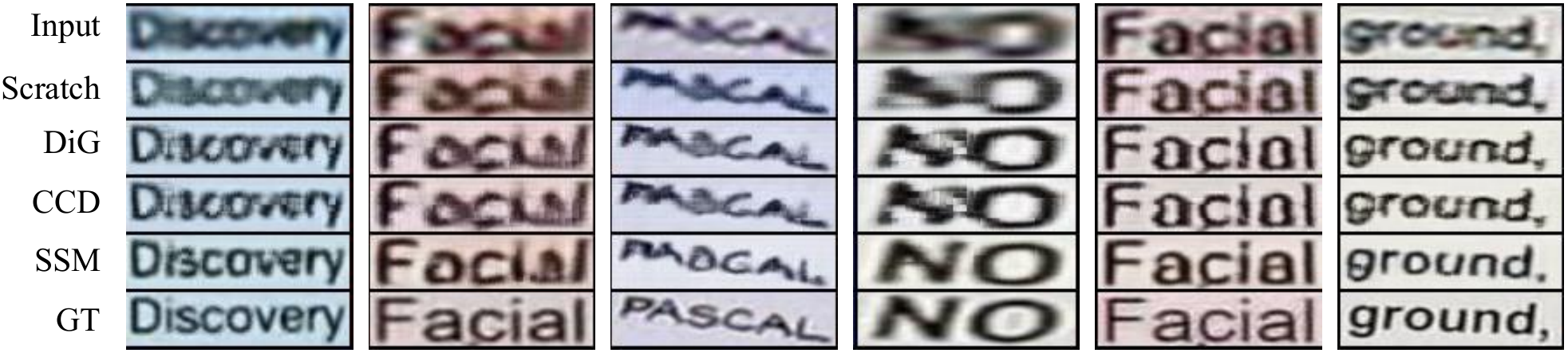}
         \caption{Text super-resolution results}
         \label{appendixfig:vis_sr}
\end{subfigure}
\caption{Illustration of (a) the text segmentation results and (b) text image super-resolution results. All menthods are pre-trained on the OCR-CC and is ViT-Small variant.}
\label{appendixfig:vis_sr_seg}
\end{figure}

Tab.~\ref{appdixtab:sr_union} also demonstrates that our SSM-S outperforms other self-supervised methods for both SSIM and PSNR metrics when pre-trained on Union14M-U and fine-tuned using the same codebase and training strategy. Some
results on the TextZoom dataset are illustrated in Fig.~\ref{appendixfig:vis_sr}. 
\begin{table}[h]
    \centering
    \begin{subtable}{\linewidth}
        \centering
\renewcommand{\arraystretch}{1.2}
\resizebox{\textwidth}{!}{
\begin{tabular}{c|cccc|cccc}
\hline
\multirow{2}{*}{Method} & \multicolumn{4}{c}{SSIM}                                              & \multicolumn{4}{c}{PSNR}                                          \\ \cline{2-9} 
                        & Easy            & Medium          & Hard            & Avg             & Easy           & Medium         & Hard           & Avg            \\ \hline
Scratch-S       & 0.8143          & 0.6288          & 0.6845          & 0.7156          & 22.90          & 19.65          & 20.45          & 21.10          \\
DiG-S\textcolor{gray}{\shortcite{yang2022reading}}           & 0.8613          & 0.6561          & 0.7215          & 0.7522          & 23.98          & 19.85          & 20.57          & 21.60          \\
CCD-S\textcolor{gray}{\shortcite{Guan_2023_ICCV}}           & 0.8822          & 0.7005          & 0.7543          & 0.7843          & 24.40          & 20.12          & 20.18          & 21.84          \\ \hline
Scratch-S$^\star$       & 0.8122          & 0.6406          & 0.6908          & 0.7197          & 23.12          & 20.14          & 20.42          & 21.32          \\
DiG-S$^\star$\textcolor{gray}{\shortcite{yang2022reading}}           & 0.8462          & 0.6516          & 0.7121          & 0.7424          & 24.31          & 20.24          & 20.66          & 21.88          \\
CCD-S$^\star$\textcolor{gray}{\shortcite{Guan_2023_ICCV}}           & 0.8418          & 0.6495          & 0.7142          & 0.7407          & 24.12          & 20.25          & 20.74          & 21.84          \\
\rowcolor{mygray}
SSM-S           & \textbf{0.8720} & \textbf{0.6696} & \textbf{0.7346} & \textbf{0.7646} & \textbf{24.90} & \textbf{20.33} & \textbf{20.88} & \textbf{22.19} \\ 
\hline
\end{tabular}
}
        \caption{All methods are pre-trained on the OCR-CC.}
        \label{appdixtab:sr_ocrcc}
    \end{subtable}
    \begin{subtable}{\linewidth}
        \centering
\renewcommand{\arraystretch}{1.2}
\resizebox{\textwidth}{!}{
\begin{tabular}{c|cccc|cccc}
\hline
\multirow{2}{*}{Method} & \multicolumn{4}{c|}{SSIM}                                             & \multicolumn{4}{c}{PSNR}                                          \\ \cline{2-9} 
                        & Easy            & Medium          & Hard            & Avg             & Easy           & Medium         & Hard           & Avg            \\ \hline
DiG-S$^\star$\textcolor{gray}{\shortcite{yang2022reading}}                   & 0.8440          & 0.6479          & 0.7085          & 0.7393          & 24.21          & 20.20          & 20.68          & 21.83          \\
CCD-S$^\star$\textcolor{gray}{\shortcite{Guan_2023_ICCV}}                   & 0.8443          & 0.6443          & 0.7072          & 0.7378          & 24.23          & 20.15          & 20.67          & 21.82          \\
\rowcolor{mygray}
SSM-S                   & \textbf{0.8685} & \textbf{0.6697} & \textbf{0.7377} & \textbf{0.7644} & \textbf{24.64} & \textbf{20.43} & \textbf{20.99} & \textbf{22.16} \\ \hline
\end{tabular}
}
        \caption{All methods are pre-trained on the Union14M-U.}
        \label{appdixtab:sr_union}
    \end{subtable}
    \caption{Comparison with other self-supervised methods on TextZoom benchmark. $^\star$ stands for our implementation based on released pre-trained weights.}
    \label{tab:mlt}
    \vspace{-12pt}
\end{table}

\subsection{Text Recognition}
\begin{table*}[t]
\centering
\renewcommand{\arraystretch}{1.2}
\setlength{\tabcolsep}{2mm}        
\resizebox{\linewidth}{!}{
\begin{tabular}{c|cccccccc|cccccc}
\toprule
\multirow{2}{*}{Method} & \multicolumn{8}{c|}{Union14M Benchmarks}                                                                                                                                        & \multicolumn{6}{c}{Other Challenge Datasets} \\ \cline{2-15} 
                        & Artistic & Contextless & Curve & General & \begin{tabular}[c]{@{}c@{}}Multi-\\ Oritented\end{tabular} & \begin{tabular}[c]{@{}c@{}}Multi-\\ Words\end{tabular} & Salient & Avg & TT    & CTW   & COCO  & ArT   & Uber  & Avg \\ \hline
Scratch-ViT-S       & 72.7     & 81.0        & 82.6  & 82.6    & 78.1                                                       & 81.7                                                   & 79.5    & 79.8 & 90.5  & 86.0  & 75.5  & 81.5  & 82.4  & 83.2 \\
MAE-ViT-S$^{\star}$ \textcolor{gray}{\cite{he2022masked}}          & 76.0     & 81.6        & 86.2  & 82.8    & 84.0                                                       & 83.5                                                   & 83.3    & 82.5 & 91.7  & 87.8  & 76.6  & 82.8  & 83.8  & 84.6 \\
DiG-ViT-S$^{\star}$ \textcolor{gray}{\cite{yang2022reading}}           & 77.4     & 82.5        & 85.9  & 83.8    & 83.5                                                       & 84.0                                                   & 84.3    & 83.0 & 91.7  & 87.2  & 77.7  & 83.4  & 84.9  & 85.0 \\
CCD-ViT-S$^{\star}$ \textcolor{gray}{\cite{Guan_2023_ICCV}}           & 73.9     & 82.2        & 83.9  & 82.6    & 81.7                                                       & 82.2                                                   & 81.7    & 81.2 & 91.3  & 87.1  & 76.1  & 82.3  & 84.4  & 84.2 \\
\rowcolor{mygray}
SSM-ViT-S           & \textbf{78.4}     & \textbf{84.7}        & \textbf{87.5}  & \textbf{84.0}    & \textbf{85.8}                                                       & \textbf{84.6}                                                   & \textbf{85.2}    & \textbf{84.3} & \textbf{92.9}  & \textbf{88.2}  & \textbf{78.1}  & \textbf{83.4}  & \textbf{86.5}  & \textbf{85.8} \\ \hline
$\uparrow$                         & \rbf{+5.7}     & \rbf{+3.7}        & \rbf{+4.9}  & \rbf{+1.4}    & \rbf{+7.7}                                                      & \rbf{+2.9}                                                   & \rbf{+5.7}    & \rbf{+4.5} & \rbf{+2.4}  & \rbf{+2.2}  & \rbf{+2.6}  & \rbf{+1.9}  & \rbf{+4.1}  & \rbf{+2.6} \\
$\triangle$                         & \gbf{+1.0}     & \gbf{+2.2}        & \gbf{+1.6}  & \gbf{+0.2}    & \gbf{+2.3}                                                       & \gbf{+0.6}                                                   & \gbf{+0.9}    & \gbf{+1.3} & \gbf{+1.2}  & \gbf{+1.0}  & \gbf{+0.4}  & \ybf{+0.0}  & \gbf{+0.6}  & \gbf{+0.8} \\ 
\bottomrule
\end{tabular}
}
\caption{Comparison with other self-supervised methods on Union14M benchmarks and other challenge datasets. $\uparrow$ and $\triangle$ represent the performance gains relative to the "train from scratch" model and the second-best model, respectively. $^{\star}$ stands for our implementation.}
\label{appendixtab:compare_ssl}
\end{table*}
\begin{table*}[t]
\centering
\renewcommand{\arraystretch}{1.2}
\setlength{\tabcolsep}{2mm}        
\resizebox{0.99\textwidth}{!}{
\begin{tabular}{cccccccccccccc}
\hline
\multicolumn{1}{l}{\multirow{2}{*}{Label Fraction}} & \multicolumn{1}{c}{\multirow{2}{*}{Method}}     & \multicolumn{3}{c}{Regular} & \multicolumn{6}{c}{Irregular}            & \multicolumn{2}{c}{Occluded} & \multicolumn{1}{c}{\multirow{2}{*}{Avg.}}  \\
\cmidrule(lr){3-5} \cmidrule(lr){6-11} \cmidrule(lr){12-13}
\multicolumn{1}{l}{}                                & \multicolumn{1}{c}{}                        & IIIT    & SVT     & IC13    & IC15 & SVTP & CUTE & COCO & CTW  & TT   & HOST & WOST                      &                       \\ \hline
\multirow{4}{*}{1\% (27.8K)}                                & \multicolumn{1}{l|}{Scratch-ViT-Small}  & 12.6  & 3.9  & 10.3 & 7.56 &  3.41 &  6.9 &  2.2 &  4.6 &  4.5 &  5.4 &  \multicolumn{1}{c|}{6.0}    & 5.2                   \\
                                                    & \multicolumn{1}{l|}{DiG-ViT-Small}      & 88.4  & 86.2 & 89.9 & 79.0 & 76.6 & 77.8 & 54.8 & 67.9 &67.2 & 33.2 & \multicolumn{1}{c|}{53.3}   & 62.9                  \\ 
                                                    & \multicolumn{1}{l|}{CCD-ViT-Small}      & 89.3  & 86.5 & 88.8 & 76.5 & 80.1 & 74.7 & 54.9 & 65.5 &67.8 & 38.4 & \multicolumn{1}{c|}{55.9}   & 63.7                  \\ 
                                                    \rowcolor{mygray}
                                                    & \multicolumn{1}{l|}{SSM-ViT-Small}      & \textbf{91.3}  & \textbf{86.7} & \textbf{91.5} & \textbf{79.9} & \textbf{80.6} & \textbf{80.9} & \textbf{57.2} & \textbf{69.6} & \textbf{70.9} & \textbf{37.5} & \multicolumn{1}{c|}{60.2}   & \textbf{66.0}                  \\ 
                                                    \hline
\multirow{4}{*}{10\% (278K)}                               & \multicolumn{1}{l|}{Scratch-ViT-Small}  & 78.4  & 73.6 & 81.8 & 66.8 & 64.8 & 56.6 &  43.2 & 48.9 & 54.4 & 30.7 & \multicolumn{1}{c|}{48.4}   & 52.3                  \\
                                                    & \multicolumn{1}{l|}{DiG-ViT-Small}      & 95.3  & 94.4 & 95.9 & 85.3 & 87.9 & 91.7 & 67.1 & 80.5 & 81.1 & 42.1 & \multicolumn{1}{c|}{64.0}   &73.5                  \\ 
                                                    & \multicolumn{1}{l|}{CCD-ViT-Small}      & 95.9  & 94.1 & 96.6 & 87.1 & 89.9 & 94.1 & 69.2 & 81.6 & 84.3 & 63.4 & \multicolumn{1}{c|}{76.2}   &78.2                  \\ 
                                                    \rowcolor{mygray}
                                                    & \multicolumn{1}{l|}{SSM-ViT-Small}      & \textbf{98.0}  & \textbf{96.8} & \textbf{97.1} & \textbf{88.4} & \textbf{90.7} & \textbf{73.3} & \textbf{87.4} & \textbf{83.9} & \textbf{87.4} &  \textbf{63.4} & \multicolumn{1}{c|}{\textbf{78.3}}   & \textbf{80.8}                  \\

                                                    \hline
\multirow{4}{*}{100\% (2.78M)}                              & \multicolumn{1}{l|}{Scratch-ViT-Small}  & 95.0  & 92.9 & 94.9 & 85.2 &  86.7 & 88.9 & 66.1 &  78.8 & 81.0 & 44.8 & \multicolumn{1}{c|}{67.9}   & 73.4                  \\
                                                    & \multicolumn{1}{l|}{DiG-ViT-Small}      & 97.7  & 96.1 & 97.3 & 88.6 & 91.6 &96.2 &75.0 & 86.3 & 88.9 & 56.0 &  \multicolumn{1}{c|}{75.7}   &80.7                  \\ 
                                                     & \multicolumn{1}{l|}{CCD-ViT-Small}      & 98.0  & 96.4 & 98.3 & 90.3 & 92.7 &98.3 &76.7 & 86.5 & 91.3 & 77.3 &  \multicolumn{1}{c|}{86.0}   &84.9                  \\ 
                                                    \rowcolor{mygray}
                                                    & \multicolumn{1}{l|}{SSM-ViT-Small}      & \textbf{98.9}  & \textbf{98.0} & \textbf{98.5} & \textbf{90.8} & \textbf{95.0} & \textbf{98.3} & \textbf{78.6} & \textbf{87.8} & \textbf{92.4} & 77.2 &  \multicolumn{1}{c|}{\textbf{86.0}}   & \textbf{86.2}                  \\ \hline
\end{tabular}
}
\caption{ Comparison results when training with different percentages of the ARD.``Scratch-$\ast$": without self-supervised pre-training.}
\label{appendixtab:data_ratio}
\vspace{-1.5em}
\end{table*}
\noindent \textbf{Comparison with CCD on Union14M benchmarks and other challenging datsets}
Due to time constraints, we do not reproduce CCD on Union14M-U during the main content submission. To provide a more comprehensive comparison, we follow CCD's settings and conducted pre-training on Union14M-U, with a total of 3 epochs and a warm-up time of 0.3 epochs. Subsequently, we performed fine-tuning on Union14M-L with 10 epochs for fair comparison. During our replication process, the loss converged to around 1.88 after three epochs of training. We confirmed with the authors that CCD also converges to around 2 in loss after pre-training on OCR-CC. On average, each epoch required approximately 8 hours of training time. The comparison results of DiG~\cite{yang2022reading}, CCD~\cite{Guan_2023_ICCV} and SSM are illustrated on Tab.~\ref{appendixtab:compare_ssl}. CCD does not achieve superior performance, as it obtains average performances of 81.2\% on Union14M benchmarks and 84.2\% on other challenging datasets, which are 3.1\% and 1.6\% lower than our SSM, respectively. We speculate that this might be attributed to the following two factors:
1) The training setting of CCD with only 3 epochs might have limited its ability to fully acquire latent knowledge from Union14M-U. However, it's worth noting that CCD's training cost for each epoch, which requires 8 hours, makes training for 20 epochs as in MAERec-S and SSM extremely expensive. (Hence, we do not have enough GPU resources and time to reproduce CCD with 20 epochs.) 2) User
In Union14M-U, there might be more diverse text images with varying font styles, complex backgrounds, and longer character sequences. CCD's use of an unsupervised segmentation algorithm based on K-means may struggle to provide accurate segmentation pseudo-labels, introducing additional noise and limiting the performance of pre-training. 
Considering that SSM does not require the preparation of pseudo-labels, and each epoch of training only takes 80 minutes, Tab.~\ref{appendixtab:compare_ssl} demonstrates that SSM is a straightforward and effective method with generality and superior performance.

\noindent \textbf{Details of fine-tuning with different data ratios} Due to space constraints, Fig. 4 in the main content does not show the evaluation datasets for the models trained at different data ratios. To ensure a fair comparison with DiG and CCD, we conducted evaluations on the benchmarks of regular text images(IIIT, SVT, IC13), irregular text images(IC15, SVTP, CUTE, COCO, CTW, TT), and occluded text images(HOST, WOST). We report the average word accuracy across these benchmarks. Tab.~\ref{appendixtab:data_ratio} shows that SSM
outperforms the previous state-of-the-art method CCD-ViT-series by 2.3\%, 2.6\%
and 1.3\%, respectively.

\begin{figure*}[t]
\centering
\includegraphics[width=\linewidth]{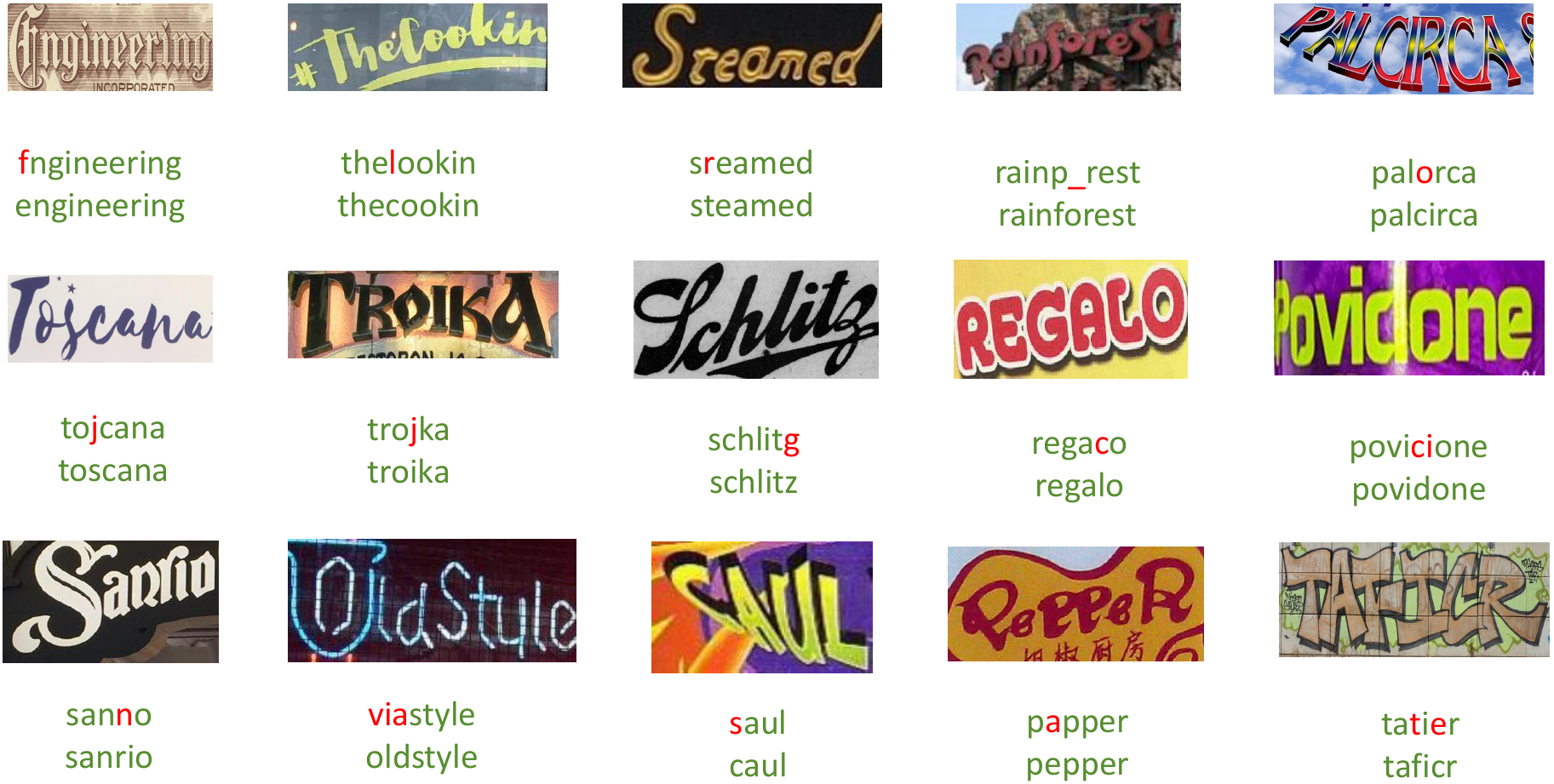}
\caption{Qualitative recognition results of \textbf{Artistic} scenes. The top and bottom strings are predicted by 'DiG-ViT-Small'(DiG-S) and 'SSM-ViT-Small'(SSM-S), with red indicating errors. }
\label{appendixfig:vis_rec_art}
\end{figure*}

\begin{figure*}[t]
\centering
\includegraphics[width=\linewidth]{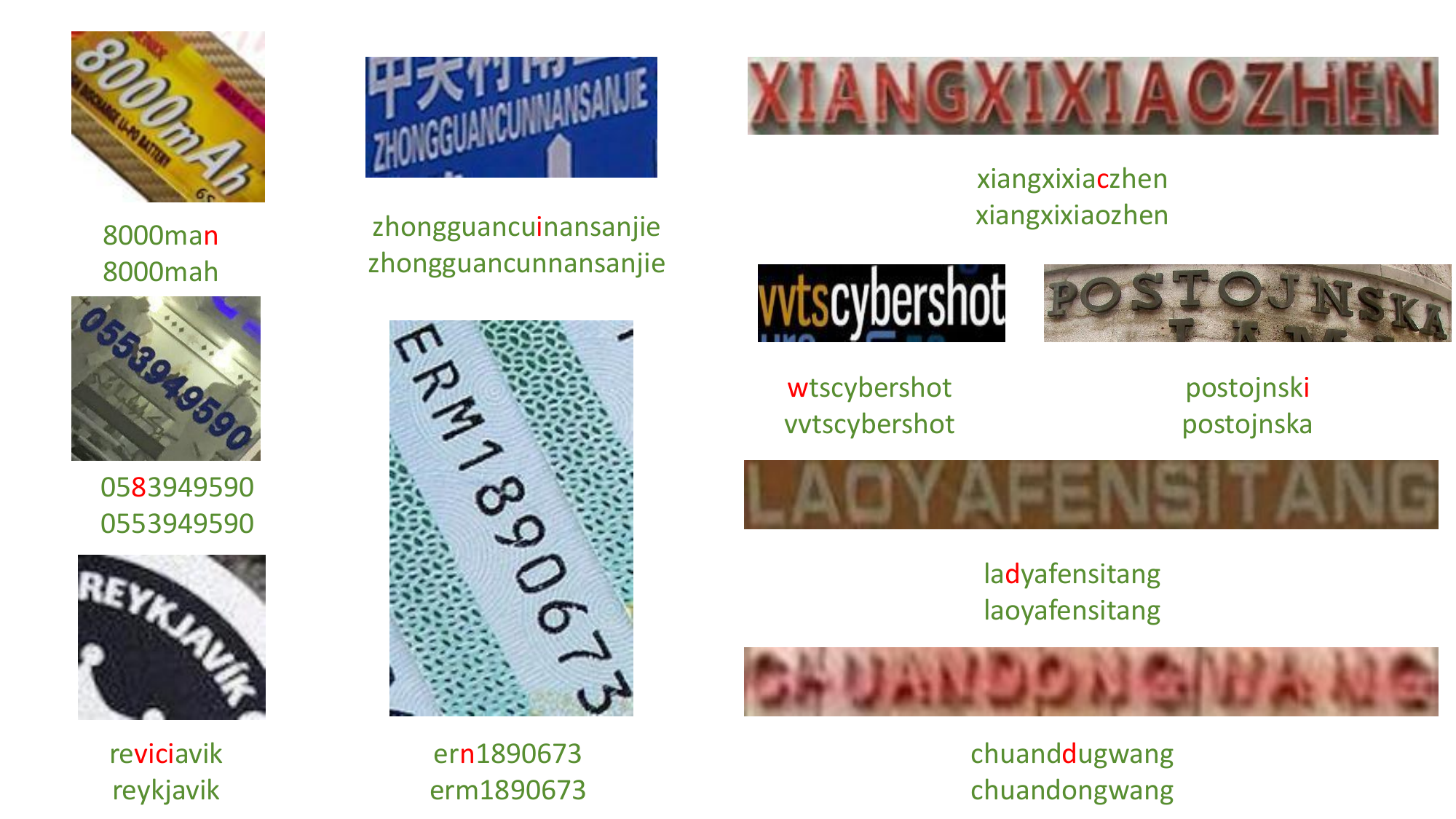}
\caption{Qualitative recognition results of \textbf{Contextless} scenes. The top and bottom strings are predicted by 'DiG-ViT-Small'(DiG-S) and 'SSM-ViT-Small'(SSM-S), with red indicating errors. }
\label{appendixfig:vis_rec_contextless}
\end{figure*}

\begin{figure*}[t]
\centering
\includegraphics[width=0.9\linewidth]{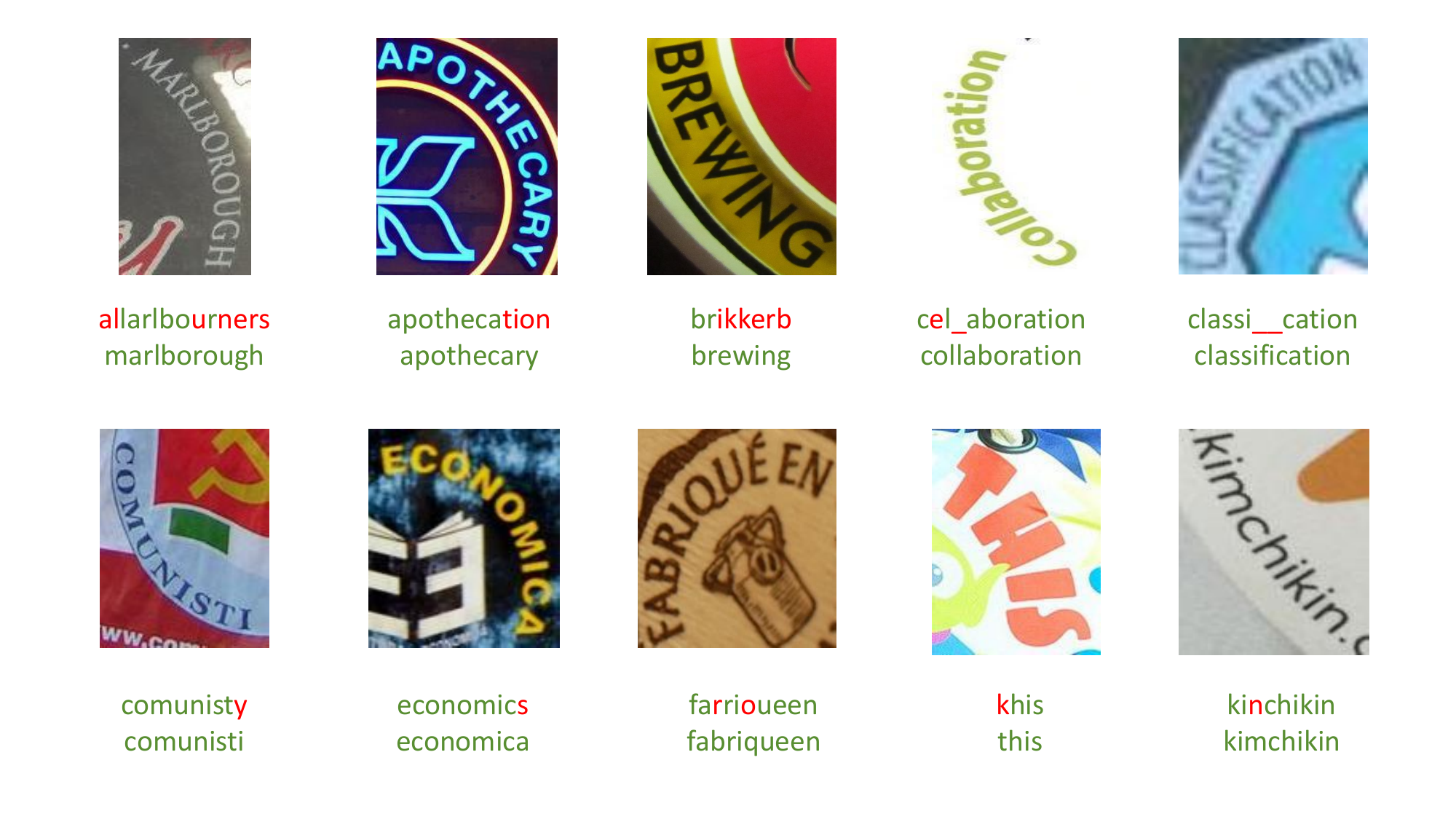}
\caption{Qualitative recognition results of \textbf{Curve} scenes. The top and bottom strings are predicted by 'DiG-ViT-Small'(DiG-S) and 'SSM-ViT-Small'(SSM-S), with red indicating errors. }
\label{appendixfig:vis_rec_curve}
\end{figure*}

\begin{figure*}[b]
\centering
\includegraphics[width=0.9\linewidth]{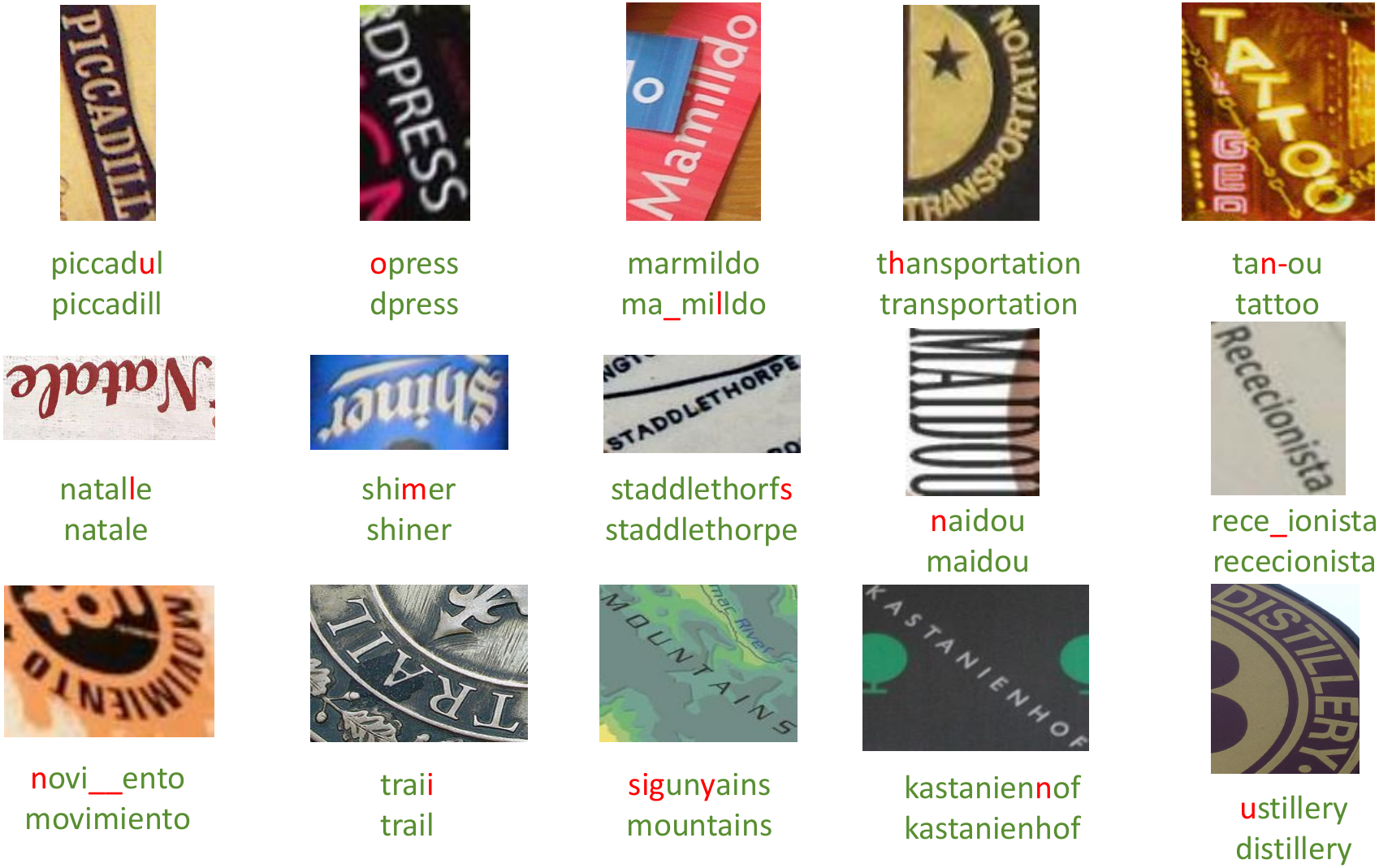}
\caption{Qualitative recognition results of \textbf{Multi-Oritented} scenes. The top and bottom strings are predicted by 'DiG-ViT-Small'(DiG-S) and 'SSM-ViT-Small'(SSM-S), with red indicating errors. }
\label{appendixfig:vis_rec_multiori}
\end{figure*}

\begin{figure*}[ht]
\centering
\includegraphics[width=\linewidth]{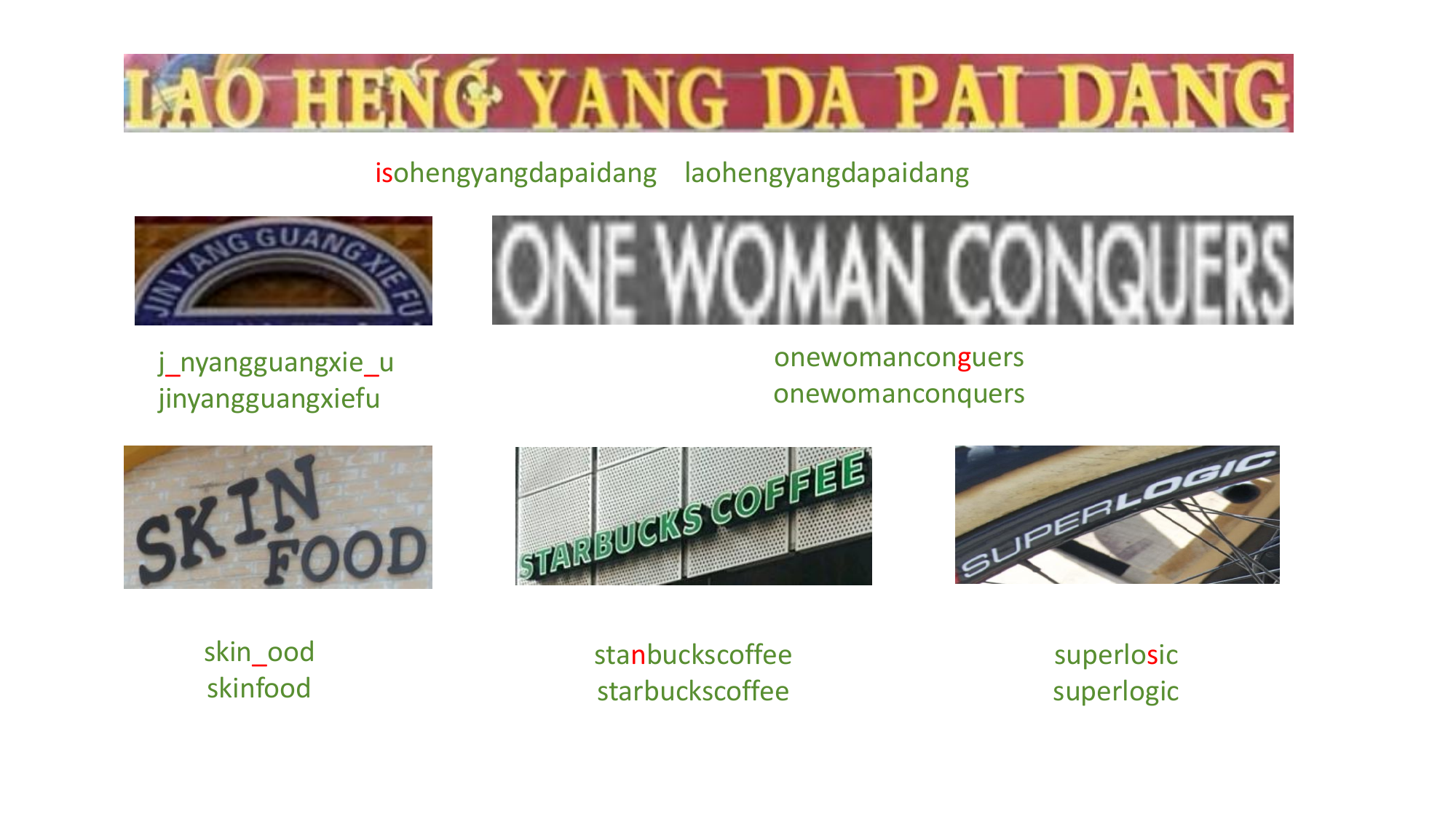}
\caption{Qualitative recognition results of \textbf{Multi-Words} scenes. The top and bottom strings are predicted by 'DiG-ViT-Small'(DiG-S) and 'SSM-ViT-Small'(SSM-S), with red indicating errors. }
\label{appendixfig:vis_rec_multiwords}
\end{figure*}

\begin{figure*}[ht]
\centering
\includegraphics[width=\linewidth]{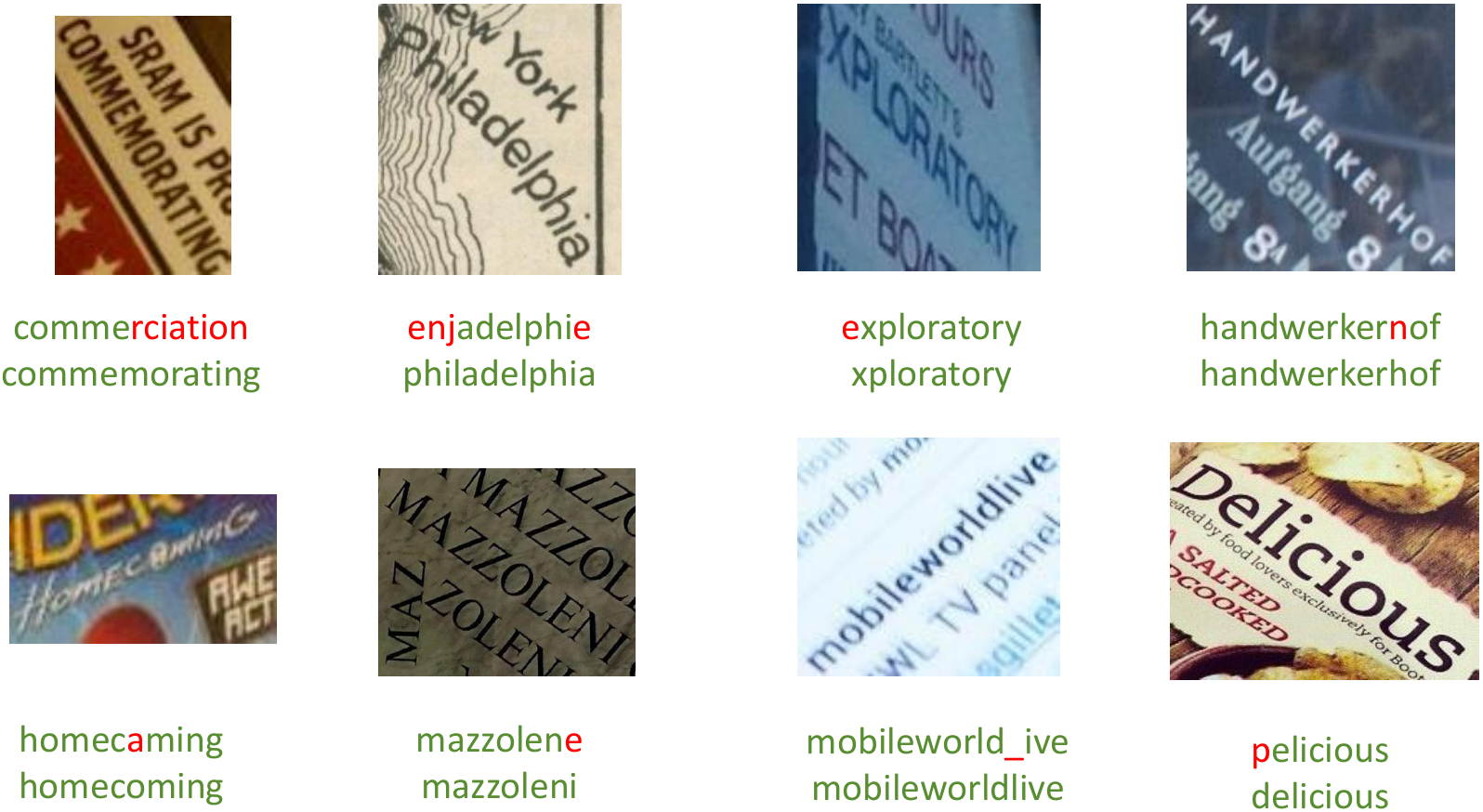}
\caption{Qualitative recognition results of \textbf{Salient} scenes. The top and bottom strings are predicted by 'DiG-ViT-Small'(DiG-S) and 'SSM-ViT-Small'(SSM-S), with red indicating errors. }
\label{appendixfig:vis_rec_salient}
\end{figure*}

\noindent \textbf{More Qualitative recognition results} In Fig.~\ref{appendixfig:vis_rec_art} to Fig.~\ref{appendixfig:vis_rec_salient}, we present additional high-quality recognition results to showcase the effectiveness and versatility of SSM in challenging scenarios. In these examples, our method demonstrates a certain level of linguistic reasoning ability to achieve robust recognition in complex backgrounds, long texts, and multi-directional scenarios.

\section{Discussion and Limitations}
\subsection{Comparasions with MIM-like methods}
Our SSM does not use masking operations to hide character pixels containing linguistic information. Despite the distortion of character shapes and linguistic information, SSM can still observe the complete global context, while MIM-like methods can only see a small portion of local character patches and cannot benefit from the complete global context. Therefore, our model emphasizes understanding the global context during the reconstruction process. 

Furthermore, the absence of masking operations makes our method more versatile for pretraining with convolutional networks or multi-scale ViT backbones without concerns about information leakage.
\subsection{About Symmetry of Superimposition}
Our superimposition input is a symmetric image in which the global content is disrupted, and all pixels of the input is visible due to no masking operations. Therefore, it is inevitable that the model relies on symmetry information during the process of reconstructing a specific direction image from symmetric overlay inputs. However, this symmetry information does not hinder the model from learning global context at both the texture-level and semantic-level. In terms of learning semantic-level global context during the feature reconstruction process, there is no issue with this symmetry information because it exists only at the pixel level. For the pixel-level reconstruction process aimed at learning texture-level global context, the symmetry information actually guides the model to focus on symmetric bidirectional positions to help the model capture global texture structures.
\subsection{Design philosophy of our architecture}
We utilize a lightweight pixel reconstruction decoder and a simple directional hint token to achieve decoupled reconstruction for inputs in different directions. This approach allows the understanding of context to be driven into the Encoder, while avoiding the dominance of the Decoder.
\subsection{Limitations}
Although our work improves the performance of several downstream tasks, including text recognition, text segmentation, and text super-resolution, it still has some limitations, which can be summarized as follows:
\begin{itemize}
    \item Using geometrically augmented input with original and flipped images in the teacher branch for feature reconstruction supervision could be an alternative. Superimposed images may introduce decoupling errors, impeding accurate reconstruction. However, using original images may encounter input misalignment issues. Unfortunately, due to time and resource constraints before the deadline, we have not fully explored the advantages and disadvantages of these two strategies.
    \item Due to a small subset of text images having skewed perspectives or curvature, our symmetric overlay reconstruction may not guarantee large overlapping areas. 
     \item For curved text with complex backgrounds or fonts, pixel-level reconstruction may not capture local details well or restore the correct viewing angle.
    \item Additionally, due to computational resources and training costs limitations, we were unable to conduct experiments on larger-scale ViT Backbones. However, the model sizes of ViT-Tiny and ViT-Small are sufficient to achieve good performance in real-world applications
\end{itemize}

\section{Visualized Reconstruction Results}
\begin{figure*}[ht]
\centering
\captionsetup[subfigure]{justification=centering}
\begin{subfigure}[b]{\linewidth}
         \centering
         \includegraphics[width=\textwidth]{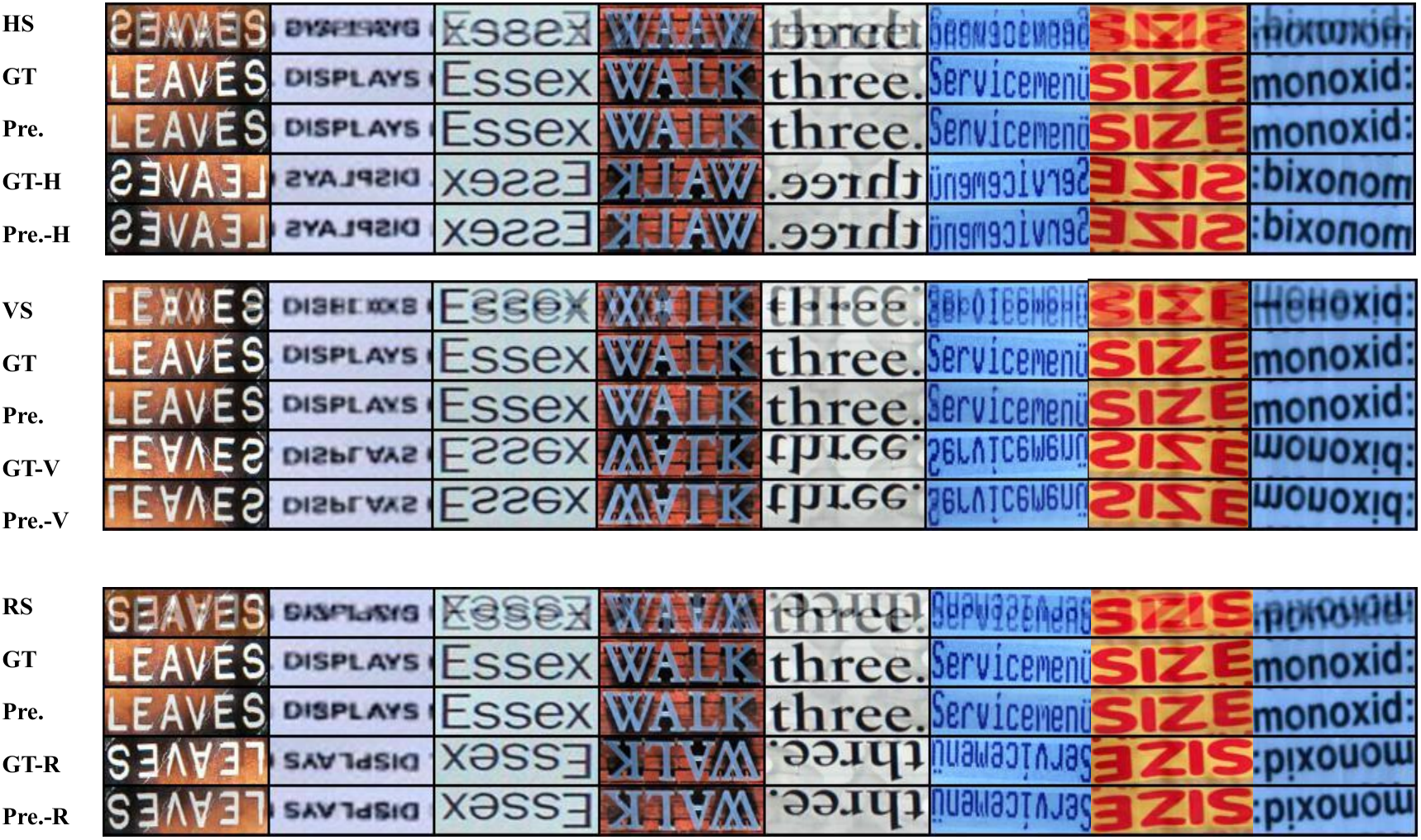}
         \caption{}
         \label{}
\end{subfigure}

\begin{subfigure}[b]{\linewidth}
         \centering
         \includegraphics[width=\textwidth]{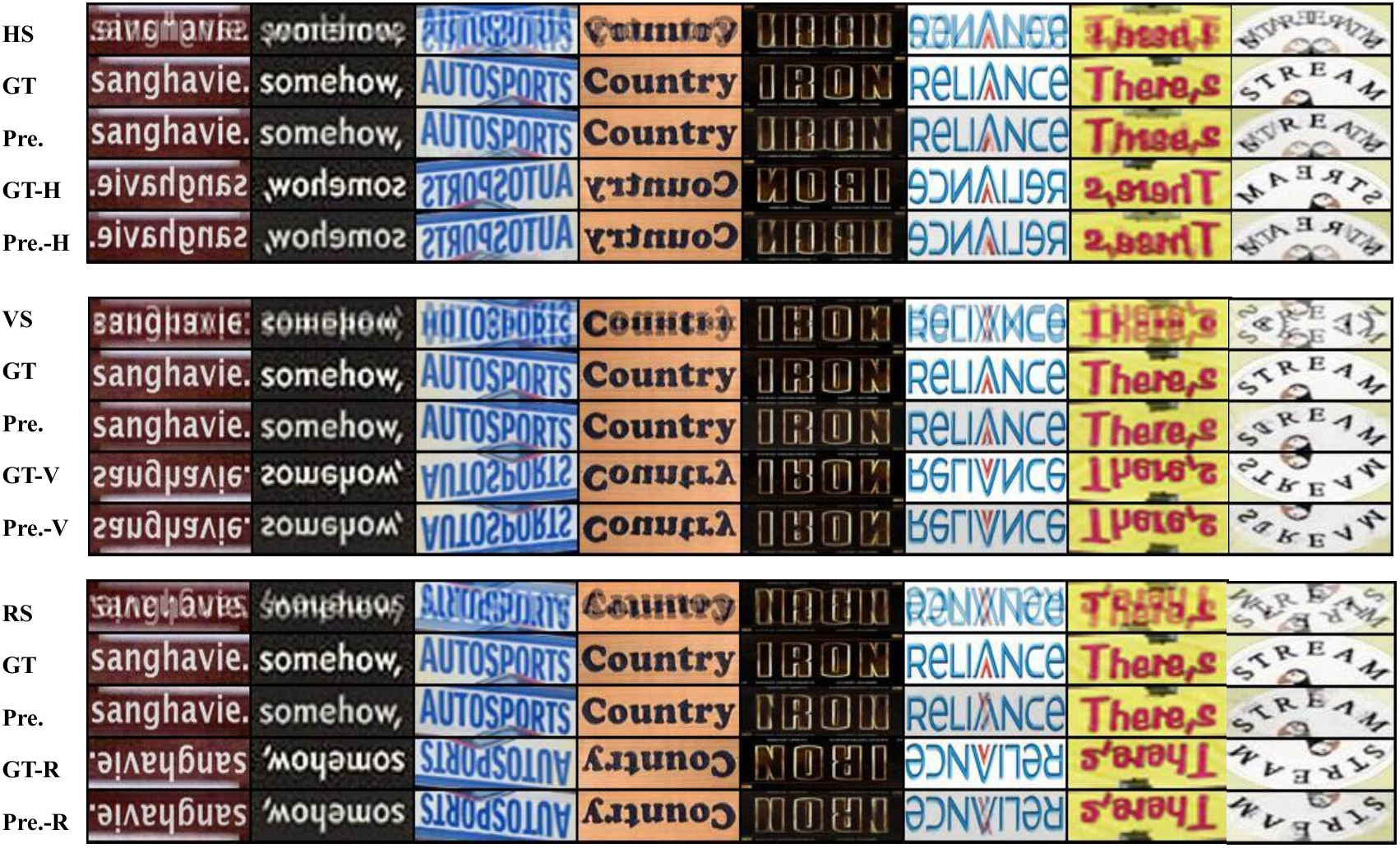}
         \caption{}
         \label{}
\end{subfigure}
\caption{Reconstruction Visualization of Regular text images. \textbf{GT}: the original image. \textbf{HS/ VS/ RS}: horizontal/ vertical / roteated superimposed input. \textbf{Pre.} indicates the pixel prediction of \textbf{GT}. \textbf{GT-H/ V/ R}: the inverted view of the \textbf{GT} (HFlip, VFlip, 180-degree rotation view, respectively). \textbf{Pre.-H/ V/ R}:the pixel prediction of \textbf{GT-H/ V/ R}. }
\label{appendixfig:recon_vis_a}
\end{figure*}

\begin{figure*}[ht]
\centering
\captionsetup[subfigure]{justification=centering}
\begin{subfigure}[b]{\linewidth}
         \centering
         \includegraphics[width=\textwidth]{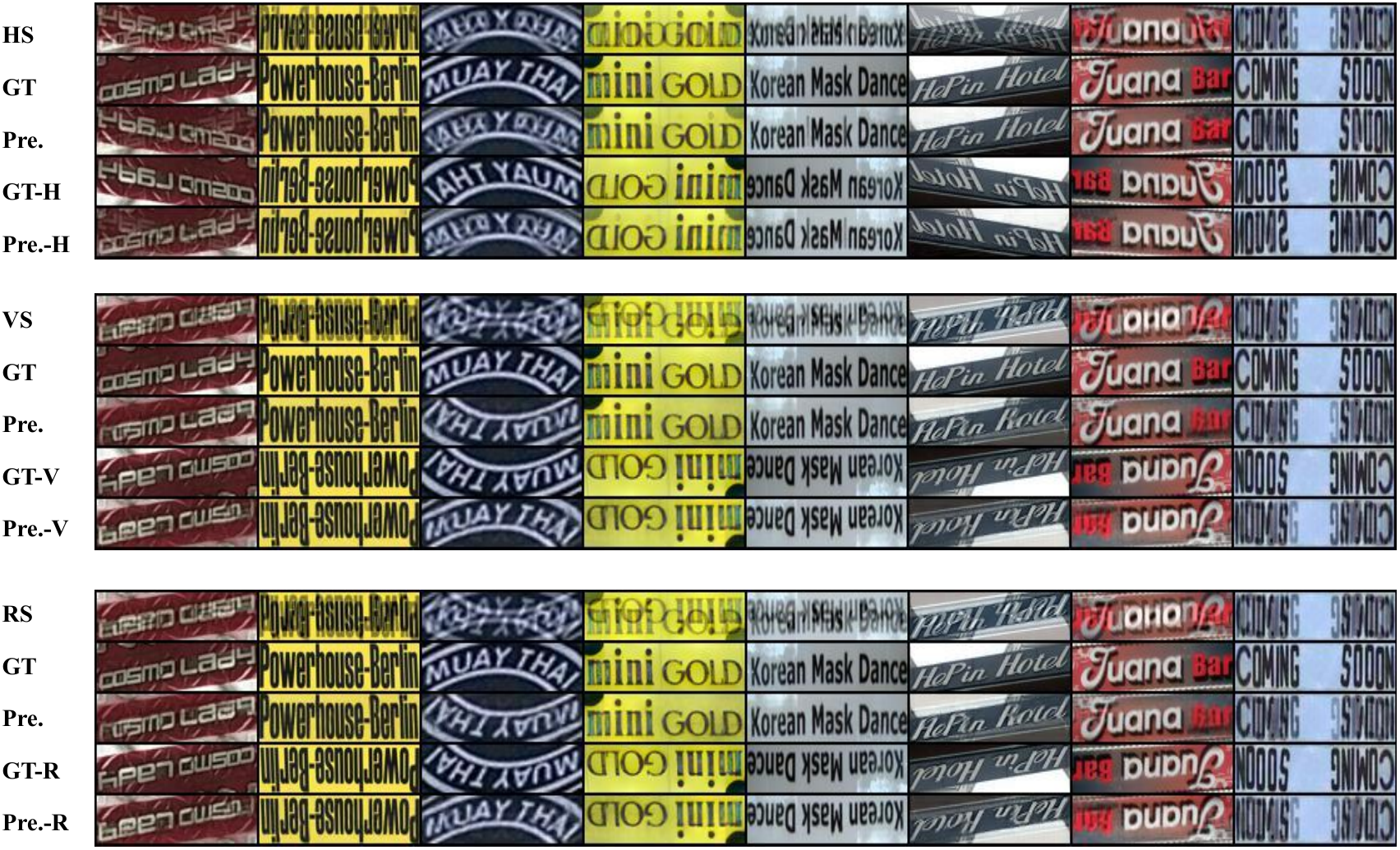}
         \caption{}
         \label{}
\end{subfigure}

\begin{subfigure}[b]{\linewidth}
         \centering
         \includegraphics[width=\textwidth]{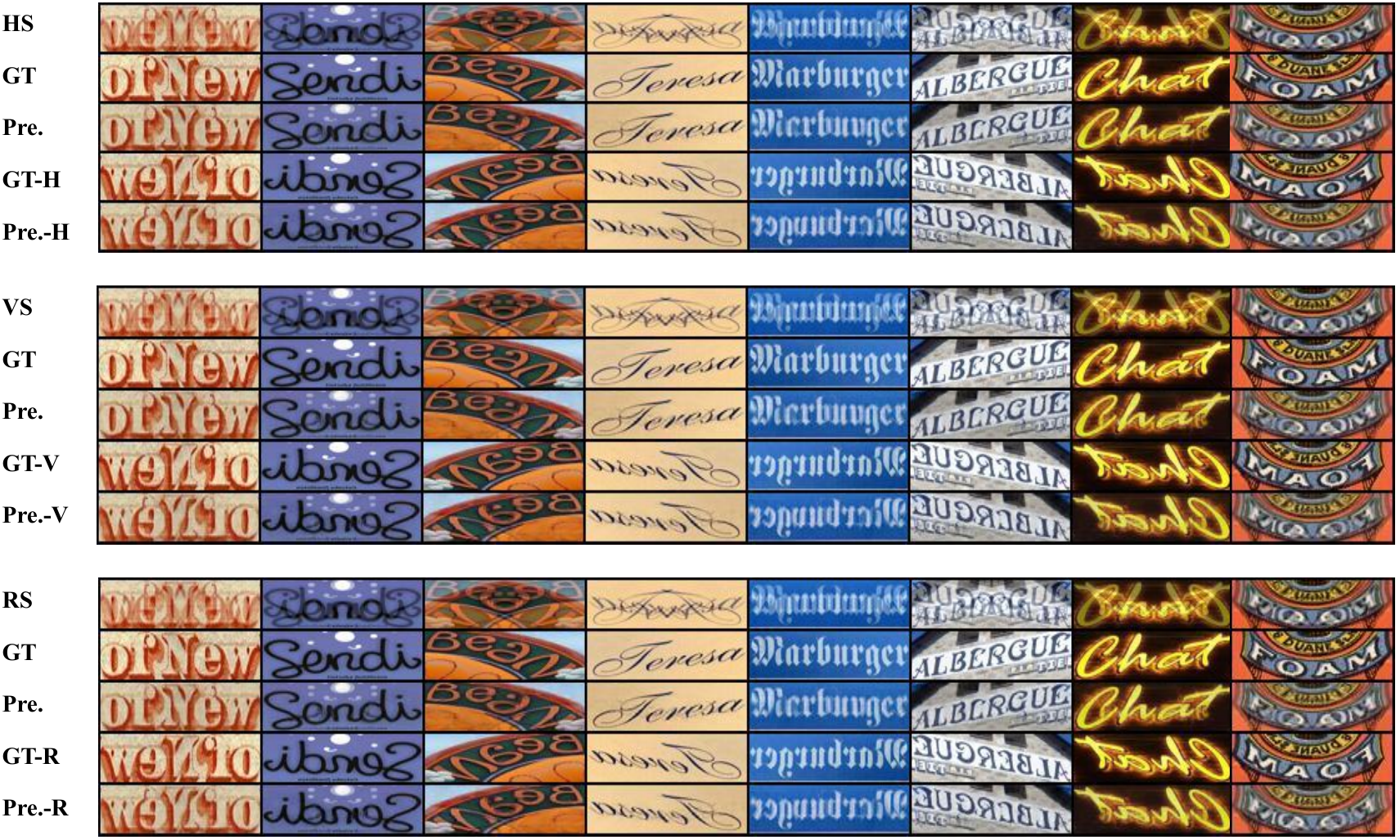}
         \caption{}
         \label{}
\end{subfigure}
\caption{Reconstruction Visualization of Multi-Words and Artistic text images. \textbf{GT}: the original image. \textbf{HS/ VS/ RS}: horizontal/ vertical / roteated superimposed input. \textbf{Pre.} indicates the pixel prediction of \textbf{GT}. \textbf{GT-H/ V/ R}: the inverted view of the \textbf{GT} (HFlip, VFlip, 180-degree rotation view, respectively). \textbf{Pre.-H/ V/ R}:the pixel prediction of \textbf{GT-H/ V/ R}. }
\label{appendixfig:recon_vis_b}
\end{figure*}

In Fig.~\ref{appendixfig:recon_vis_a} and Fig.~\ref{appendixfig:recon_vis_b}, we show more reconstruction visualization images in pixel level with various text images of evaluation datasets. We found good results with most images, including horizontal, curved, and skewed text images. However, there is poor local recovery and an inability to decompose text pixels in different directions for a small number of curved or complex background images.


\end{document}